\documentclass[10pt,twocolumn,letterpaper]{article}

\usepackage[pagenumbers]{cvpr} 

\usepackage{graphicx}
\usepackage{amsmath}
\usepackage{amssymb}
\usepackage{booktabs}
\usepackage{bm}
\usepackage{bbm}
\usepackage{multirow}
\usepackage[table]{xcolor}
\usepackage{enumitem}

\usepackage[pagebackref,breaklinks,colorlinks]{hyperref}

\usepackage[capitalize]{cleveref}
\crefname{section}{Sec.}{Secs.}
\Crefname{section}{Section}{Sections}
\Crefname{table}{Table}{Tables}
\crefname{table}{Tab.}{Tabs.}

\def\cvprPaperID{10116} 
\def\confName{CVPR}
\def\confYear{2023}

\begin{document}

\title{
	Selectively Hard Negative Mining for Alleviating Gradient Vanishing \\
	in Image-Text Matching}


\author{Zheng Li$^{1}$, Caili Guo$^{1}$, Xin Wang$^{1}$, Zerun Feng$^{1}$, Zhongtian Du$^{2}$ \\
	$^1$Beijing University of Posts and Telecommunications \\
	$^2$China Telecom Digital Intelligence Technology Co., Ltd. \\
}
\maketitle

\begin{abstract}
Recently, a series of Image-Text Matching (ITM) methods achieve impressive performance.
However, we observe that most existing ITM models suffer from gradients vanishing at the beginning of training, which makes these models prone to falling into local minima.
Most ITM models adopt triplet loss with Hard Negative mining (HN) as the optimization objective.
We find that optimizing an ITM model using only the hard negative samples can easily lead to gradient vanishing.
In this paper, we derive the condition under which the gradient vanishes during training.
When the difference between the positive pair similarity and the negative pair similarity is close to 0, the gradients on both the image and text encoders will approach 0.
To alleviate the gradient vanishing problem, we propose a Selectively Hard Negative Mining (SelHN) strategy, which chooses whether to mine hard negative samples according to the gradient vanishing condition.
SelHN can be plug-and-play applied to existing ITM models to give them better training behavior.
To further ensure the back-propagation of gradients, we construct a Residual Visual Semantic Embedding model with SelHN, denoted as RVSE++.
Extensive experiments on two ITM benchmarks demonstrate the strength of RVSE++, achieving state-of-the-art performance. 
\end{abstract}
\vspace{-10pt}

\section{Introduction}
Image-Text Matching (ITM)~\cite{li2019visual, chen2020imram, liu2020graph} aims to establish the correspondence between image and text, which is fundamental to various vision-language understanding tasks.
The challenge of ITM is the heterogeneous gap between image and text.
The ITM model needs to accurately learn the semantic correspondence between image and text.
Existing ITM methods generally follow a common framework, which contains a two-branch neural network as image and text encoders with a triplet loss~\cite{schroff2015facenet} as the optimization objective.
The triplet consists of three parts: anchor, positive and negative.
We take each sample in the training set as an anchor.
The samples semantically related to the anchor are called positive samples, and the irrelevant samples to the anchor are called negative samples.
The anchor and the corresponding positive sample form a positive pair. 
Similarly, the anchor and the negative sample form a negative pair.
The mainstream approach to implementing ITM is to learn a model that measures a matching score between image and text.
We denote the matching scores of positive and negative pairs as $ s_{p} $ and $ s_{n} $, respectively.
The triplet loss can guide the model to achieve $ s_{p} > s_{n} $.

\begin{figure}
	\centering
	\includegraphics[width=0.65\linewidth]{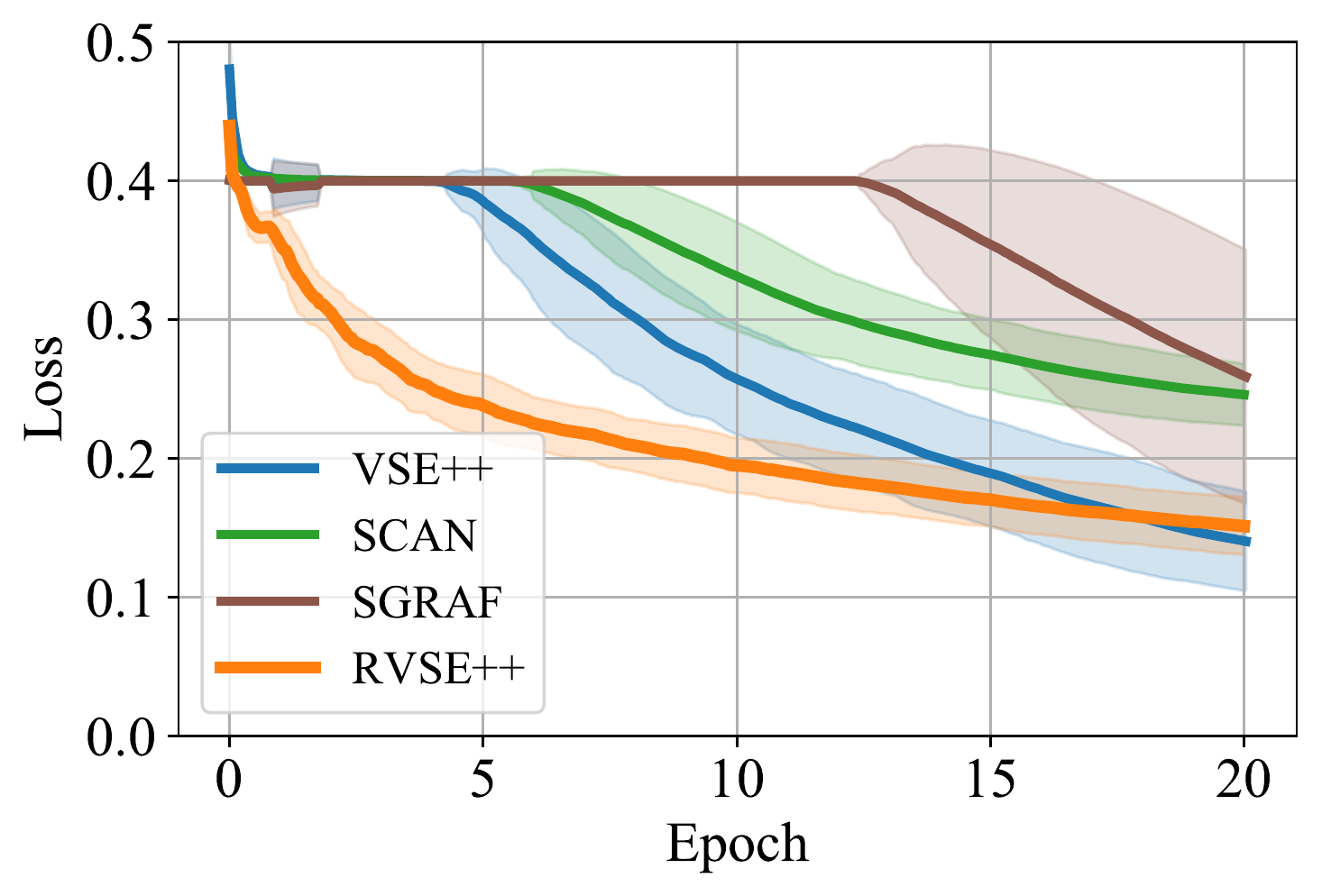}
	\vspace{-5pt}
	\caption{
		Plotting the losses against training epochs on the Flickr30K dataset.
		The solid line represents the smoothed mean of the loss, and the shaded area represents its standard deviation.
	}
	\vspace{-15pt}
	\label{train_loss}
\end{figure}
ITM methods can be mainly divided into two categories, \textit{i.e.}, the Visual Semantic Embedding (VSE) methods, such as VSE++~\cite{faghri2018vse++}, and the Cross-Attention (CA) methods, such as SCAN~\cite{lee2018stacked}.
The VSE method embeds the whole image and sentence into a joint embedding space.
The matching score can be calculated by a simple similarity metric.
The CA methods obtain the matching score by calculating the cross-attention between the image regions and words.
There are also some methods that combine VSE and CA methods for global and local matching, such as SGRAF~\cite{diao2021similarity}.
For the optimization objective of the ITM model, Faghri \textit{et al.}~\cite{faghri2018vse++} propose triplet loss with Hard Negative mining (HN).
HN mines the hard negative samples with the largest $ s_{n} $ in each batch, which yields significant gains in matching performance.
Most existing ITM methods~\cite{lee2018stacked, diao2021similarity, zhang2022negative} adopt triplet loss with HN as the optimization objective.

Recently, a series of ITM methods achieve impressive performance.
However, we observe that most ITM models have bad training behavior.
As shown in \figurename~\ref{train_loss}, we plot the losses against training epochs on the Flickr30K dataset~\cite{young2014image} for several classical ITM models VSE++, SCAN, and SGRAF.
They all adopt the triplet loss with HN as the optimization objective.
The three models have a period in which the loss does not decrease at the beginning of training. 
Especially for SGRAF with the most complex model structure, the loss of SGRAF does not decrease in the first 12 epochs.
In this paper, we experimentally observe that the reason for the bad training behavior is that the ITM model suffers from gradient vanishing, which makes the model prone to falling into local minima. As a result, the model cannot fully utilize its representational powers.

In Deep Metric Learning (DML), some studies~\cite{schroff2015facenet, xuan2020hard} focus on the bad training behavior caused by training with triplet loss.
They attribute the bad training behavior to optimizing hard negative samples.
Schroff \textit{et al.}~\cite{schroff2015facenet} define negative samples satisfying $ s_{n}>s_{p} $ as hard negative samples.
Various work shows that optimizing with the hard negative samples leads to bad local minima in the early stage of training~\cite{schroff2015facenet, oh2016deep, xuan2020hard}.
Some negative mining strategies~\cite{schroff2015facenet, xuan2020hard}, such as Semi-Hard Negative mining (SHN)~\cite{schroff2015facenet}, are proposed to alleviate this problem. 
These methods generally abandon optimization for triplet consisting of hard negative samples. 
But hard negative samples are crucial for learning a discriminative model~\cite{xuan2020hard}. 

In this paper, we provide a solution to the gradient vanishing during ITM model training.
Through the gradient analysis of the triplet loss with HN, we get the condition under which the gradient vanishes.
When $ (s_{n} - s_{p}) $ is close to 0, the gradients on the image and text encoders will approach 0.
To alleviate the gradient vanishing, we propose a Selectively Hard Negative Mining (SelHN) strategy.
When the condition is satisfied, SelHN uses all negative samples for optimization; when not satisfied, SelHN mines hard negative samples for training.
SelHN can be plug-and-play applied to existing ITM models to give them better training behavior.
To further ensure the back-propagation of gradients, we construct a Residual Visual Semantic Embedding model with SelHN, denoted as RVSE++.
Our main contributions are summarized as follows:
\begin{itemize}[leftmargin=*]
	\item 
	We deduce the condition for gradient vanishing during ITM model training, which can guide the design of the hard negative mining strategy.
	\item 
	We propose a SelHN strategy to alleviate gradient vanishing.
	SelHN can be plug-and-play applied to existing ITM models to give them better training behavior.
	\item 
	We constructed an RVSE++ model. 
	Extensive experiments on two ITM benchmarks demonstrate the strength of RVSE++, achieving state-of-the-art performance. 
\end{itemize}

\section{Related Work}

\subsection{Image-Text Matching Models}
ITM aims to establish the correspondence between image and text, which is fundamental to various vision and language understanding tasks, such as vision-language retrieval~\cite{goenka2022fashionvlp, lu2022cots}, image/video captioning~\cite{li2022comprehending, seo2022end}, and visual question answering~\cite{cascante2022simvqa, gupta2022swapmix}.
Existing ITM methods can be mainly divided into two categories, \textit{i.e.}, the Visual Semantic Embedding (VSE) methods~\cite{li2019visual, wang2020consensus, chen2021learning}, and the Cross-Attention (CA) methods~\cite{chen2020imram, liu2020graph, zhang2020context}.

\textbf{Visual Semantic Embedding.}
The VSE method embeds the whole image and sentence into a joint embedding space~\cite{zheng2020dual, yan2021discrete, li2022multi}..
The matching score between the image and text can be calculated by a simple similarity metric (\textit{e.g.} cosine similarity).
Therefore, the VSE method has a faster inference speed, which is suitable for real application scenarios.
Frome \textit{et al.}~\cite{frome2013devise} propose the first VSE model DeViSE, which employs the CNN and Skip-Gram~\cite{mikolov2013efficient} to project the image and sentence into a joint embedding space.
Faghri \textit{et al.}~\cite{faghri2018vse++} propose VSE++, which uses CNN to encode the image and Gated Recurrent Unit (GRU)~\cite{cho2014learning} to encode the sentence, and incorporates hard negative mining in the triplet loss. 
Chen \textit{et al.}~\cite{chen2021learning} propose a Generalized Pooling Operator (GPO) that learns the best pooling strategy to generate global embeddings.
Our proposed RVSE++ belongs to the VSE method.

\textbf{Cross-Attention.}
The CA methods obtain the matching score by calculating the cross-attention between the image regions and words~\cite{chen2020imram, liu2020graph, zhang2020context, xu2020cross, qu2021dynamic, wang2020cross}. 
Lee \textit{et al.}~\cite{lee2018stacked} propose SCAN, which measures image-text similarity by aligning image regions and words.
The cross-attention method can match more fine-grained image-text correspondence, but its matching speed is slow.
There is a part of the work that combines the VSE method and the CA method.
Diao \textit{et al.}~\cite{diao2021similarity} propose SGRAF, which contains a similarity graph reasoning and attention filtration network for local and global alignments.
Although these ITM methods achieve impressive performance by learning rich embedding representations.
We observe that most existing ITM models suffer from gradients vanishing during training.

\subsection{Loss functions for Image-Text Matching}
In recent years, a variety of loss functions~\cite{faghri2018vse++, wei2020universal, chen2020adaptive} are proposed for ITM. 
A hinge-based triplet loss~\cite{frome2013devise} is widely used as an objective to force positive pairs to have higher matching scores than negative pairs by a margin. 
Faghri \textit{et al.}~\cite{faghri2018vse++} propose triplet loss with HN, which incorporates hard negatives in the triplet loss, which yields significant gains in matching performance. 
Most state-of-the-art ITM methods~\cite{lee2018stacked, li2019visual, diao2021similarity, zhang2022negative} adopt triplet loss with HN as the optimization objective.
Some studies propose more complex sample mining~\cite{chen2020adaptive} or sample weighting~\cite{wei2020universal, wei2021universal, wei2021meta} strategies, which further improve the matching performance of ITM models. 
These methods all focus on constructing an optimization objective to learn discriminative feature representations, ignoring the gradient vanishing problem that is prone to occur in neural network training.
The SelHN proposed in this paper can outperform these complex sample weighting or mining strategies in matching performance through selectively hard negative mining.

Some research work~\cite{schroff2015facenet, xuan2020hard} in DML focus on the bad training behavior of hard negative mining.
SHN~\cite{schroff2015facenet} does not optimize hard negative samples and only mines negative samples with $ s_{n}<s_{p} $ to optimize triplet loss.
Xuan \textit{et al.}~\cite{xuan2020hard} propose Selectively Contrastive Triplet loss (SCT), which uses the contrastive loss~\cite{hadsell2006dimensionality} to optimize hard negative samples to avoid bad training behavior, and uses triplet loss to optimize the remaining samples.
These methods improve the training behavior of triplet loss, but abandon optimization for triplet consisting of hard negative samples. 
Hard negative samples are crucial for learning a discriminative model~\cite{xuan2020hard}. 
This paper deduce the condition for gradient vanishing during ITM model training.
The proposed SelHN chooses whether to mine hard negative samples according to the condition.
SelHN not only improves the training behavior of the ITM model, but also takes full advantage of hard negative mining.

\section{Methodology}
\subsection{Preliminaries}
\textbf{Image-Text Matching.}
We first introduce the related background of ITM. 
Given a set of images $ \bm{V} $ and a corresponding set of sentences $ \bm{T} $. 
Let $ V_{i} $ be an image, $ T_{i} $ be a sentence, and $ {(V_{i}, T_{i})} $ be an image-text pair.
We denote a positive pair by $ (V_{i}, T_{i}) $ and a negative pair by $ (V_{i}, T_{j})_{i \neq j} $.
Existing ITM methods generally follow a common framework, which contains a two-branch neural network as image and text encoders with a triplet loss as the optimization objective.
We denote the image encoder and text encoder as $ \bm{f}_{vis}(\cdot) $ and $ \bm{f}_{text}(\cdot) $, respectively.
The core idea behind the approach is that there exists a mapping function $ s(V_{i}, T_{i}; \bm{W}) = \bm{f}_{vis}(V_{i})^{\top} \bm{W} \bm{f}_{text}(T_{i}) $ to measure the matching score between the visual features $ \bm{f}_{vis}(V_{i}) $ and the text features $ \bm{f}_{text}(T_{i}) $, where $ \bm{W} $ denotes the parameters of $ s(\cdot, \cdot) $.
For the VSE methods, $ s(\cdot, \cdot) $ is a simple similarity metric (\textit{e.g.} cosine similarity).
For the CA method, $ s(\cdot, \cdot) $ is an attention-based matching score generator.

\begin{figure}
	\centering
	\begin{subfigure}{0.32\linewidth}
		\includegraphics[width=\linewidth]{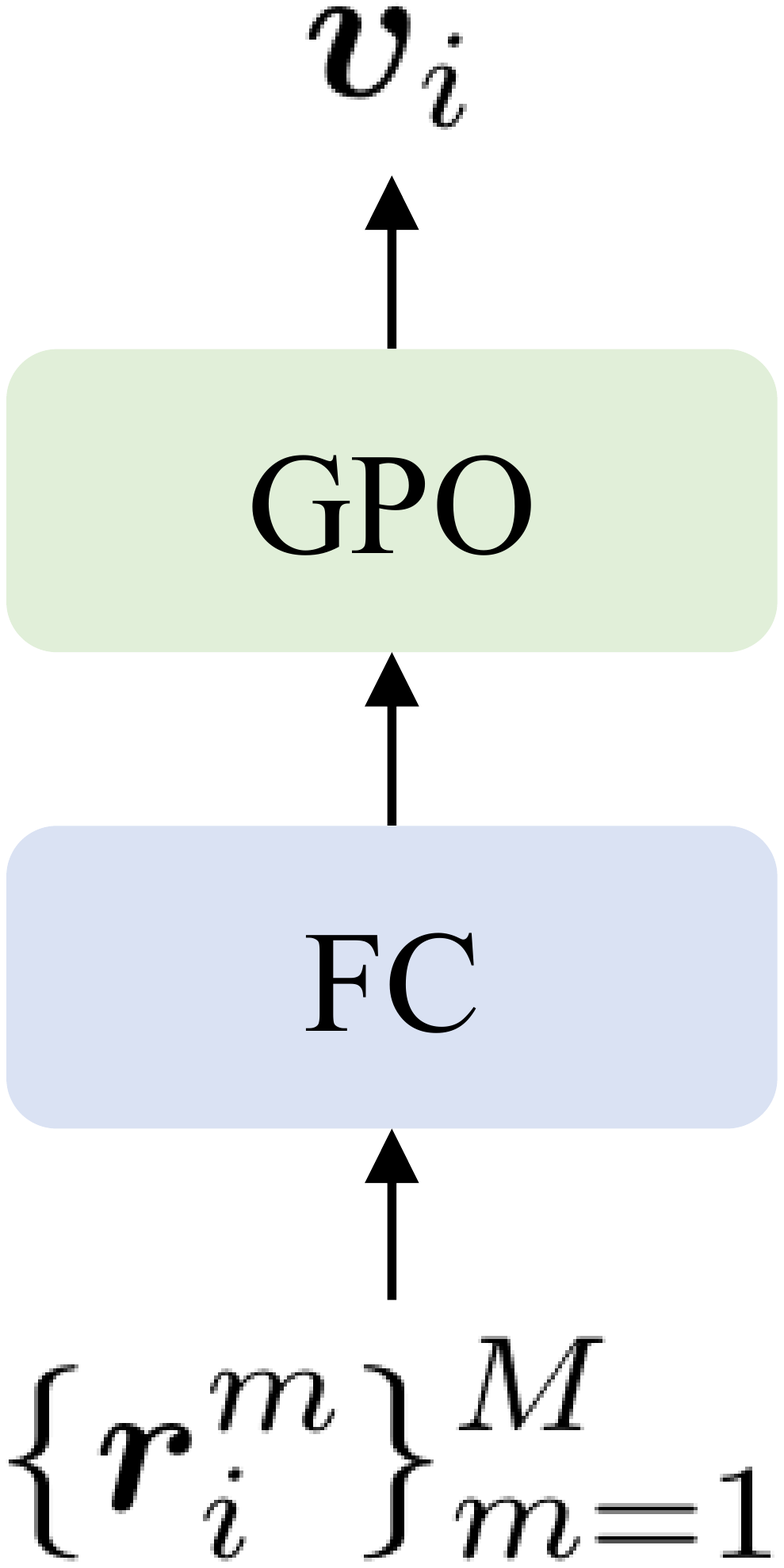}
		\caption{VSE (FC)}
		\label{vse_fc}
	\end{subfigure}
	\hfill
	\begin{subfigure}{0.32\linewidth}
		\includegraphics[width=\linewidth]{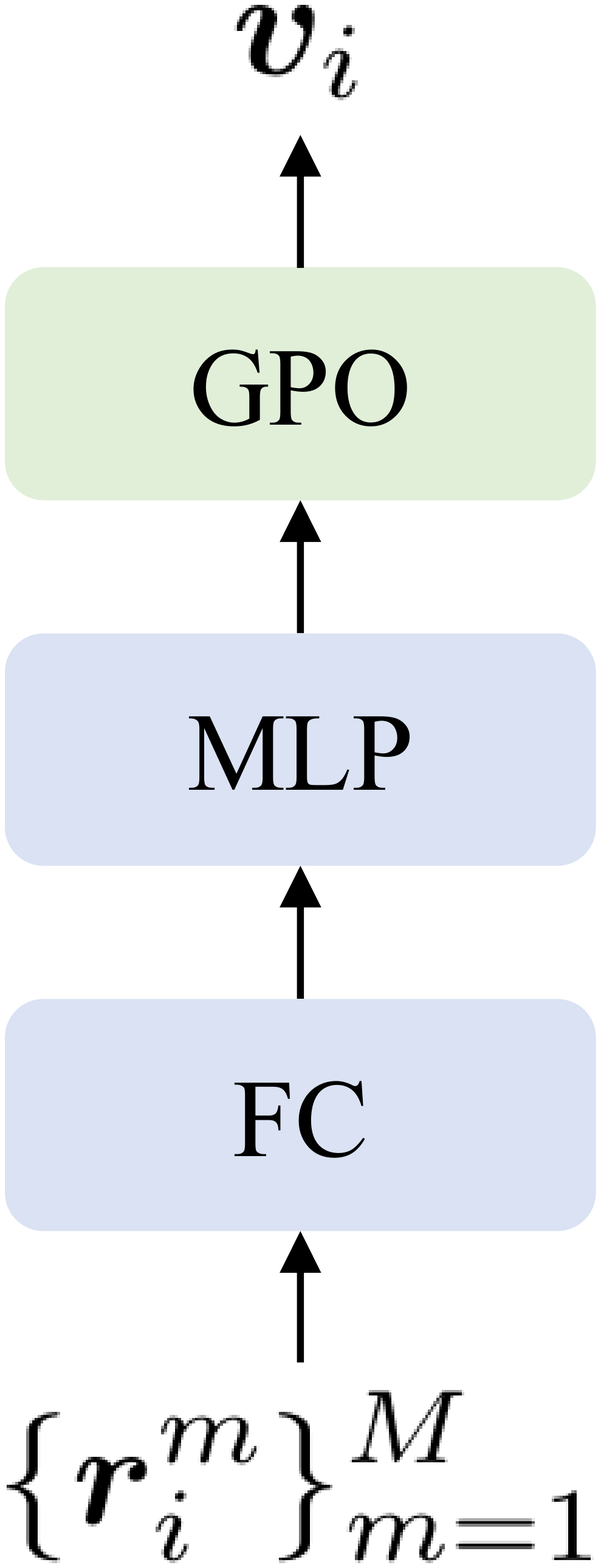}
		\caption{VSE (MLP)}
		\label{vse_mlp}
	\end{subfigure}
	\hfill
	\begin{subfigure}{0.32\linewidth}
		\includegraphics[width=\linewidth]{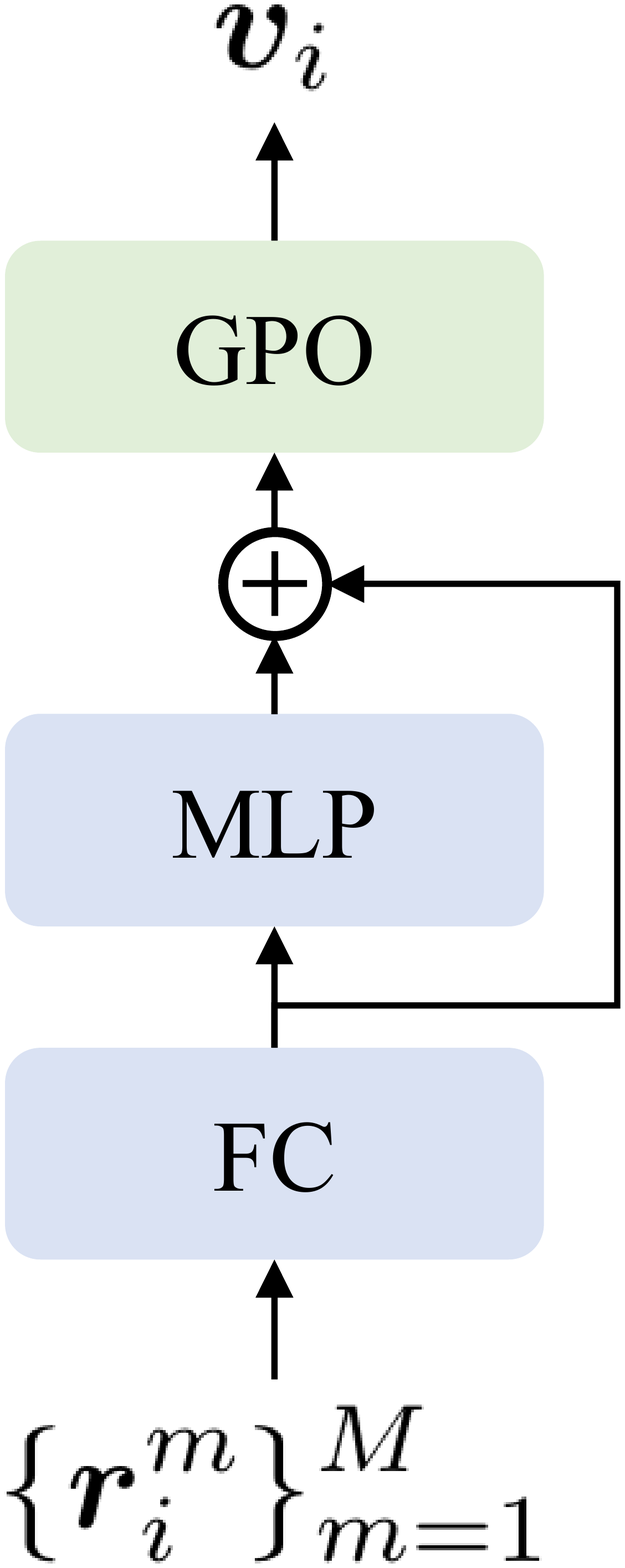}
		\caption{RVSE (MLP)}
		\label{rvse_mlp}
	\end{subfigure}
	\vspace{-5pt}
	\caption{Several image encoder architectures.}
	\vspace{-10pt}
	\label{model}
\end{figure}
\textbf{Model Architecture.}
To facilitate the experimental analysis of the gradient vanishing in ITM, we construct three simple VSE models.

\textbf{VSE (FC).}
For image encoding, we adopt the Bottom-Up and Top-Down (BUTD) attention region features~\cite{anderson2018bottom} used by most ITM methods~\cite{lee2018stacked, li2019visual}.
Given an image $ V_{i} $, we utilize Faster-RCNN~\cite{ren2015faster} to obtain $ M $ region-level features $ \{ \bm{r}_{i}^{m} \}_{m=1}^{M} $.
Then we use a Fully Connected (FC) layer to map the dimension of $ \{ \bm{r}_{i}^{m} \}_{m=1}^{M} $ to $ d $, where $ d $ is the embedding dimension. 
Finally, we use a GPO~\cite{chen2021learning} to aggregate the image feature set into the image embedding $ \bm{v}_{i} \in \mathbb{R}^{d} $.
The image encoder architecture of VSE (FC) is shown in \figurename~\ref{vse_fc}.
For text encoding, given a sentence $ T_{i} $, we adopt a Bi-GRU~\cite{bahdanau2015neural} to obtain $ N $ word-level features $ \{ \bm{w}_{i}^{n} \}_{n=1}^{N} $.
Then we also use a GPO to aggregate the text feature set into the text embedding $ \bm{t}_{i} \in \mathbb{R}^{d} $.

\textbf{VSE (MLP).}
To enhance the representation ability of the image encoder, we add a two-layer MultiLayer Perceptron (MLP) between FC and GPO based on VSE (FC), as shown in \figurename~\ref{vse_mlp}.
Except for the image encoder, all model architectures of VSE (MLP) are the same as VSE (FC).

\textbf{RVSE (MLP).}
Residual connection is an effective way to alleviate the gradient vanishing in the training of neural networks. 
Therefore, we add a residual connection between the FC and GPO based on VSE (MLP), denoted as RVSE (MLP), as shown in \figurename~\ref{rvse_mlp}.
	
For these three VSE models, we denote the trainable parameters of the image and text encoder by $ \bm{\theta}_{vis} $ and $ \bm{\theta}_{text} $, respectively. 
$ \bm{v}_{i} $ and $ \bm{t}_{i} $ can be expressed as:
\begin{equation}
	\bm{v}_{i} 
	= \bm{f}_{vis}
	\left(
	V_{i}; \bm{\theta}_{vis}
	\right),
	\bm{t}_{i} 
	= \bm{f}_{text}
	\left(
	T_{i}; \bm{\theta}_{text}
	\right),
\end{equation}
Note that both $ \bm{v}_{i} $ and $ \bm{t}_{i} $ are $ L_{2} $-normalized unit vectors.
Finally, we use cosine similarity to calculate the match score between the image and text: 
$ s(V_{i}, T_{i}) = \bm{v}_{i}^{\top} \bm{t}_{i} $.

\textbf{Loss Function.}
For subsequent experiments and gradient analysis, we revisit two classical loss functions.

\textbf{Triplet loss}~\cite{schroff2015facenet} is a classical loss function in the DML, which is widely used in face recognition~\cite{schroff2015facenet}, image retrieval~\cite{oh2016deep}, \textit{etc}.
The triplet loss applied to ITM contains both image-to-text and text-to-image directions and can be expressed as:
\begin{equation} \label{triplet}
	\begin{aligned}
		\mathcal{L}_{\text{Triplet}}
		= \sum_{i=1}^{B}
		\sum_{j=1, i \neq j}^{B}
		\left(
		[ 
		s(V_{i}, T_{j}) - s(V_{i}, T_{i}) + \lambda 
		]_{+}
		\right. \\
		+ \left. [ 
		s(V_{j}, T_{i}) - s(V_{i}, T_{i}) + \lambda 
		]_{+}
		\right),
	\end{aligned}
\end{equation}
where $ B $ is the batch size, $ \lambda $ is the margin for better similarity separation, and $ [x]_{+} = \max (x, 0) $. 

\textbf{Triplet loss with Hard Negative mining (HN)}~\cite{faghri2018vse++} incorporates the hard negative in triplet loss, which yields significant gains in matching performance.
Most state-of-the-art ITM methods~\cite{lee2018stacked, li2019visual, diao2021similarity, zhang2022negative} adopt triplet loss with HN as the optimization objective, which takes the form of:
\begin{equation} \label{HN}
	\begin{aligned}
		\mathcal{L}_{\text{T-HN}}
		= \sum_{i=1}^{B}
		\left(
		[ 
		s(V_{i}, \hat{T}_{i}) - s(V_{i}, T_{i}) + \lambda 
		]_{+}
		\right. \\
		\left. 
		+ [ 
		s(\hat{V}_{i}, T_{i}) - s(V_{i}, T_{i}) + \lambda 
		]_{+}
		\right),
	\end{aligned}
\end{equation}
where 
$ \hat{T}_{i} 
= \arg\max_{j = 1, i \neq j}^{B} 
s{(V_{i}, T_{j})} $
and
$ \hat{V}_{i} 
= \arg\max_{j = 1, i \neq j}^{B} 
s{(V_{j}, T_{i})} $ 
are the hard negative samples. 

\begin{figure}
	\centering
	\begin{subfigure}{0.49\linewidth}
		\includegraphics[width=\linewidth]{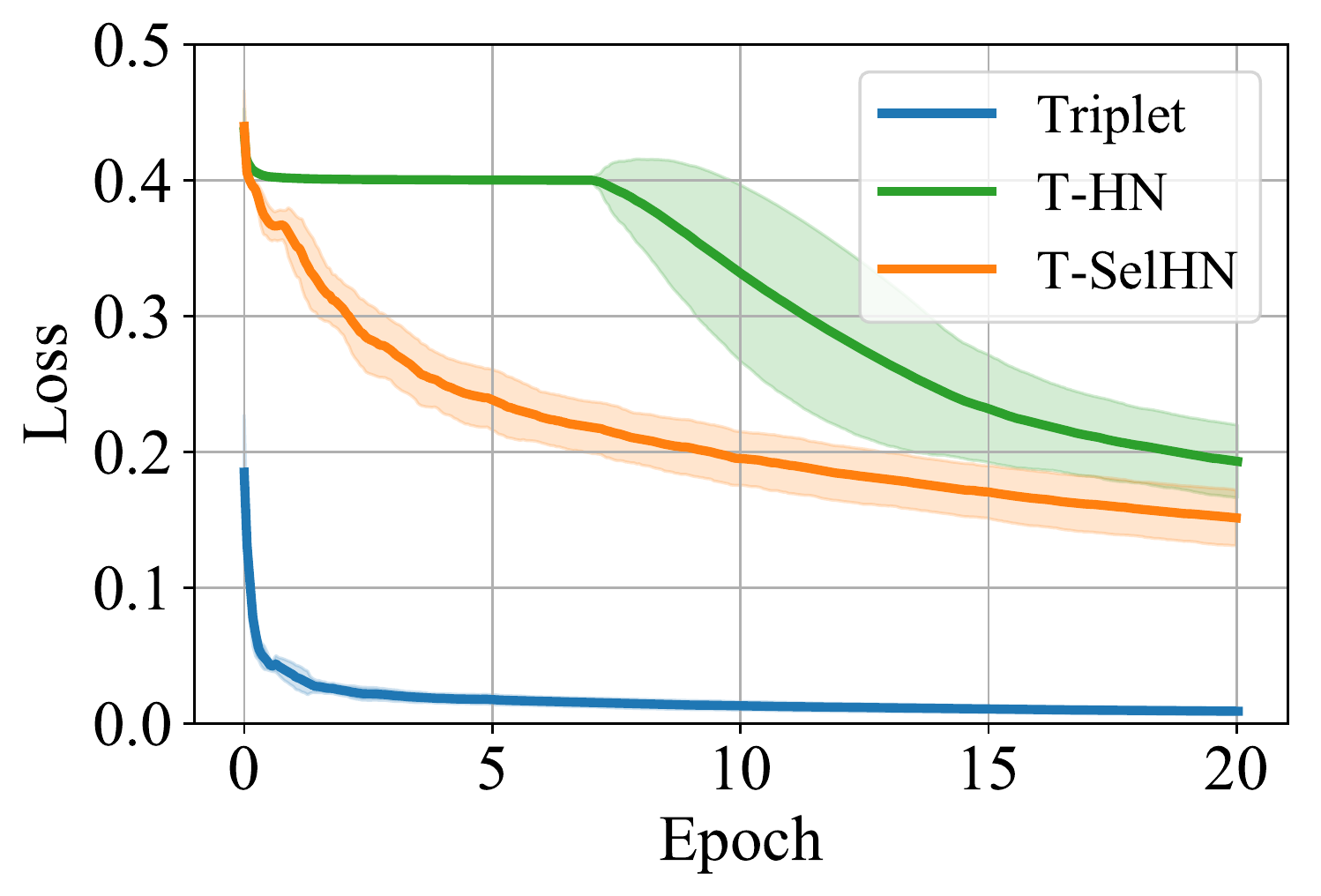}
		\caption{Loss}
		\label{train_loss_fc}
	\end{subfigure}
	\hfill
	\begin{subfigure}{0.49\linewidth}
		\includegraphics[width=\linewidth]{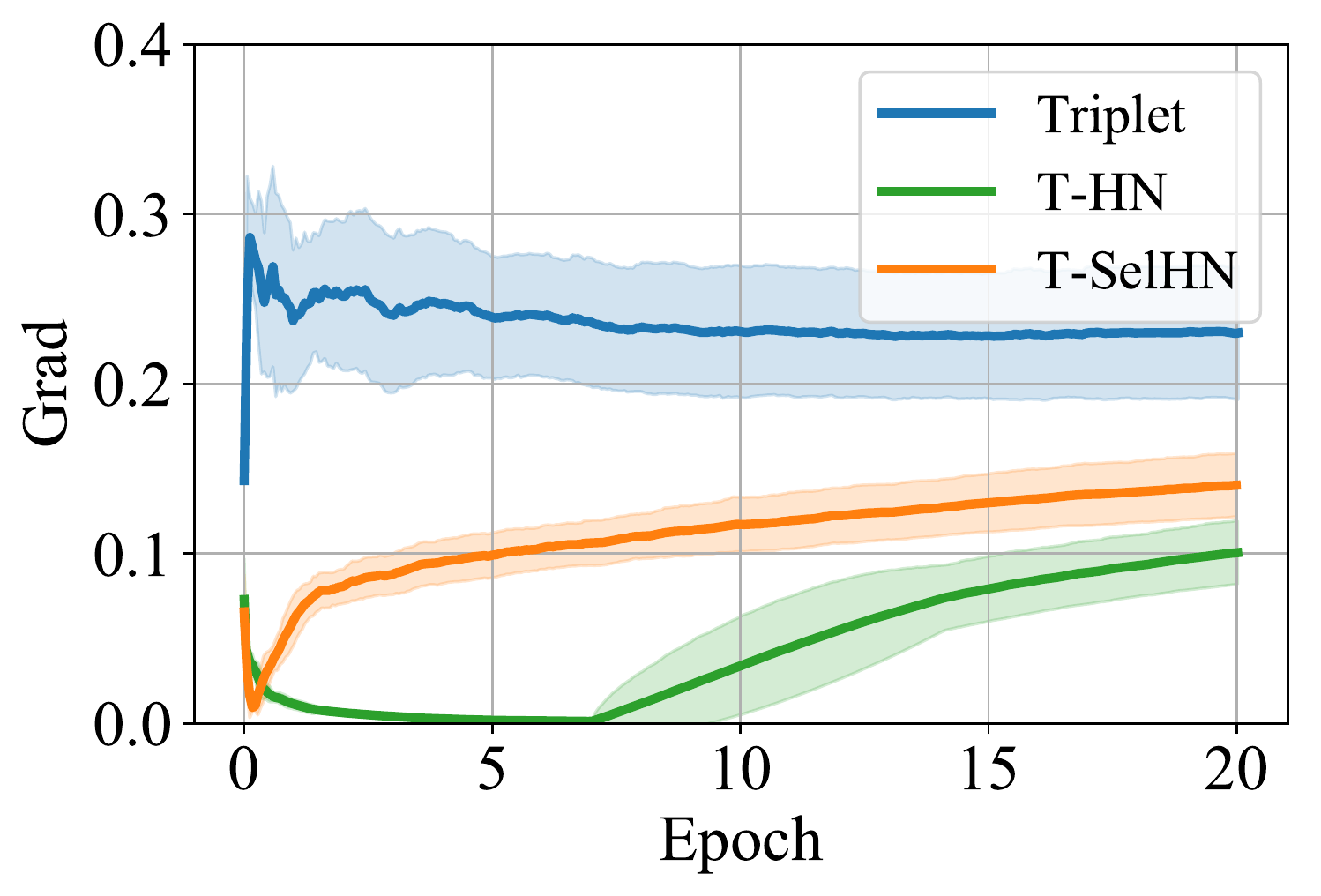}
		\caption{Gradient}
		\label{train_grad_fc}
	\end{subfigure}
	\hfill
	\begin{subfigure}{0.49\linewidth}
		\includegraphics[width=\linewidth]{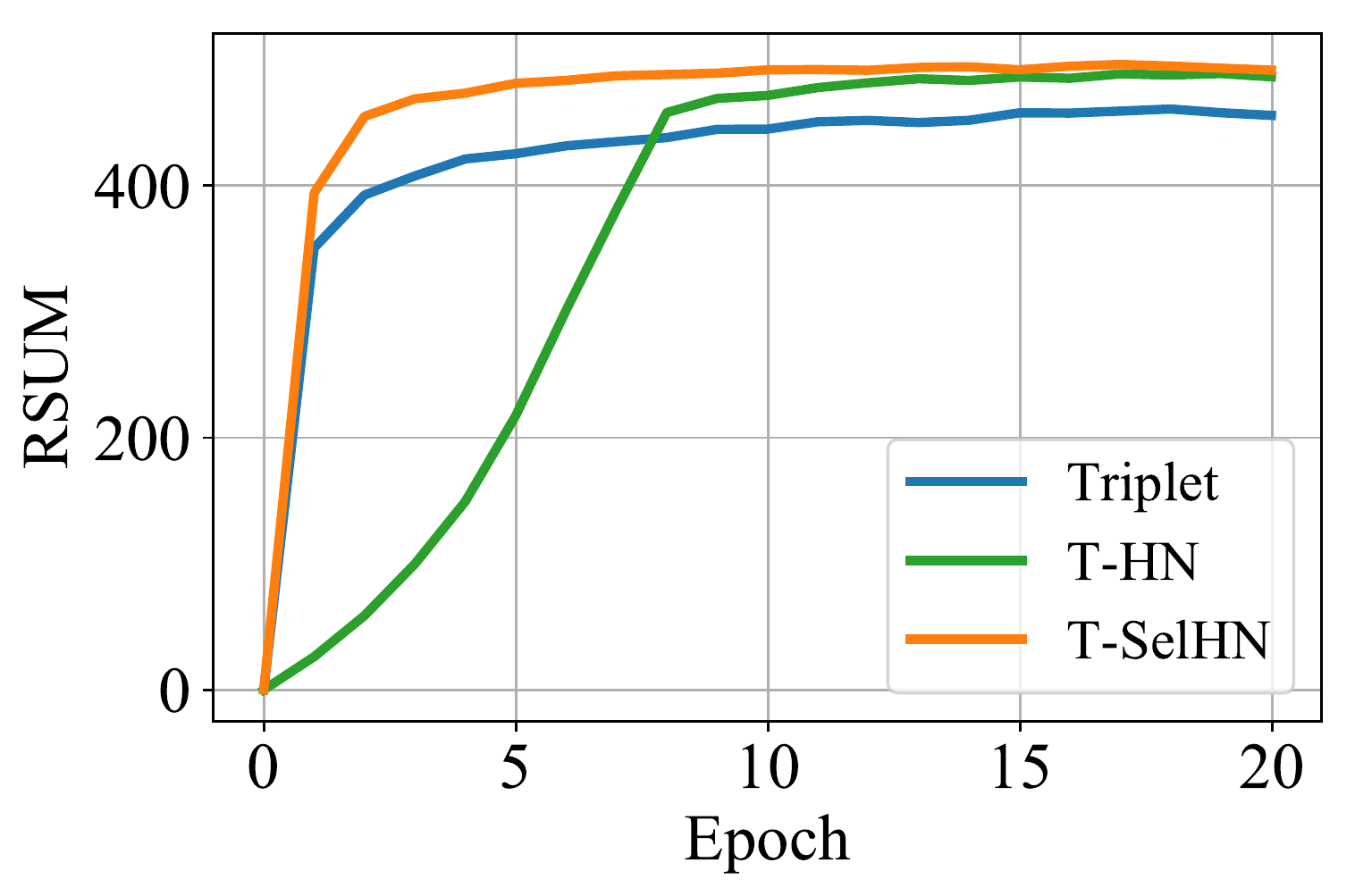}
		\caption{Sum of Recall}
		\label{train_rsum_fc}
	\end{subfigure}
	\hfill
	\begin{subfigure}{0.49\linewidth}
		\includegraphics[width=\linewidth]{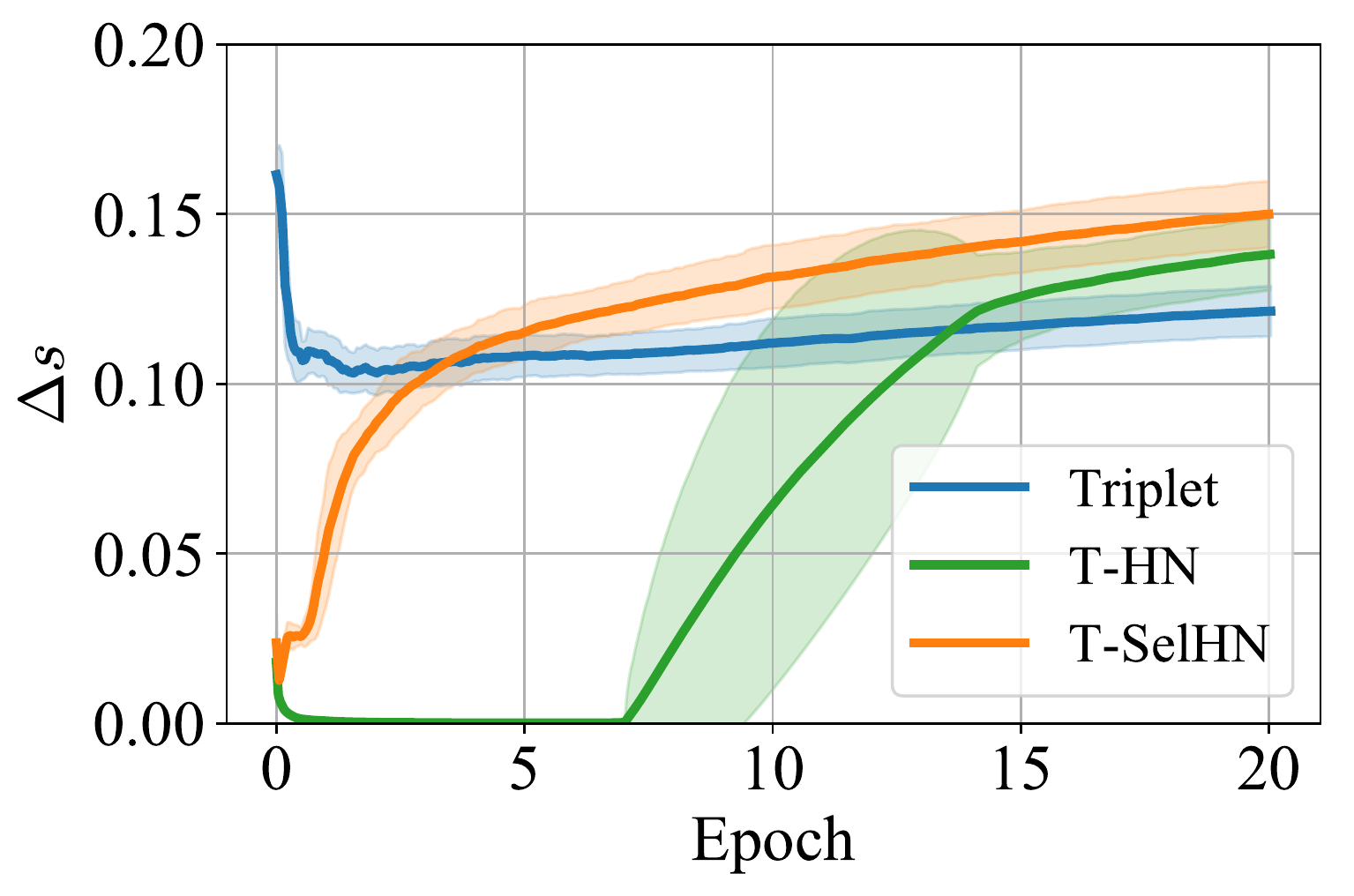}
		\caption{$ \Delta s $}
		\label{train_diff_fc}
	\end{subfigure}
	\vspace{-5pt}
	\caption{Training performance of VSE (FC) on Flickr30K.}
	\vspace{-10pt}
	\label{fc}
\end{figure}
\textbf{Model Training.}
As shown in \figurename~\ref{train_loss}, the existing ITM models generally have bad training behavior. 
To explore the reason, we use VSE (FC), VSE (MLP) and RVSE (MLP) for experimental analysis.
We train the three models on Flickr30K dataset with $ \mathcal{L}_{\text{Triplet}} $ and $ \mathcal{L}_{\text{T-HN}} $ as optimization objective. 
These models are trained using AdamW~\cite{loshchilov2018decoupled} for 20 epochs, with a batch size of 128.
The embedding dimension $ d $ is set to 1,024.
The learning rate of the model is set as 0.0005. 
The margin $ \lambda $ is set as 0.2.
The losses against epochs in training of VSE (FC) are shown in \figurename~\ref{train_loss_fc}.
It can be seen that $ \mathcal{L}_{\text{Triplet}} $ decreases steadily during training without bad training behavior.
But $ \mathcal{L}_{\text{T-HN}} $ does not decrease at 0-7 epoch. 
We count the gradient of the FC layer of the VSE (FC) model during training, as shown in \figurename~\ref{train_grad_fc}.
When using $ \mathcal{L}_{\text{T-HN}} $ training, the gradient of the FC layer is close to $ 0 $ at 0-7 epoch. 
But training with $ \mathcal{L}_{\text{Triplet}} $ does not suffer from gradient vanishing.
It is consistent with the decreasing trend of loss during training.
Therefore, we infer that the reason for the bad training behavior is the gradient vanishing, which makes the model unable to fully exploit its representational power.
Even if $ \mathcal{L}_{\text{T-HN}} $ has bad training behavior, as shown in \figurename~\ref{train_rsum_fc}, hard negative mining can improve the matching performance of the model, so most ITM models still use $ \mathcal{L}_{\text{T-HN}} $ as the optimization objective.

\begin{figure}
	\centering
	\begin{subfigure}{0.32\linewidth}
		\includegraphics[width=\linewidth]{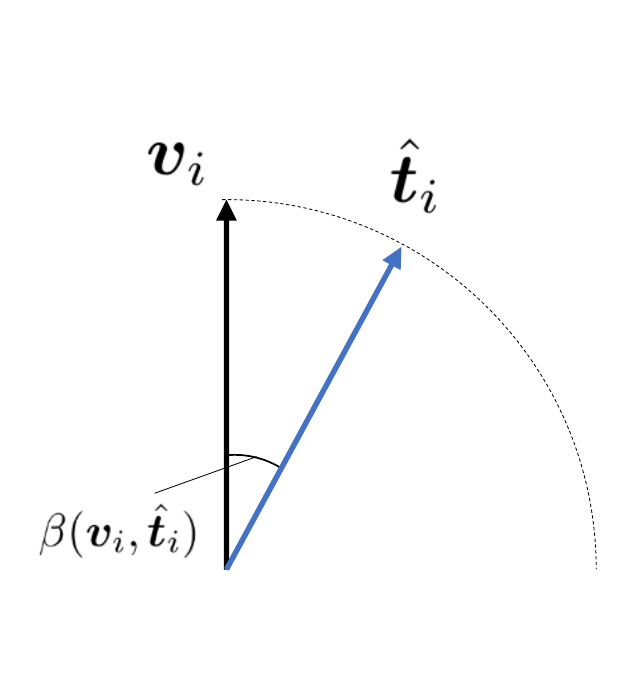}
		\caption{}
		\label{grad1}
	\end{subfigure}
	\hfill
	\begin{subfigure}{0.32\linewidth}
		\includegraphics[width=\linewidth]{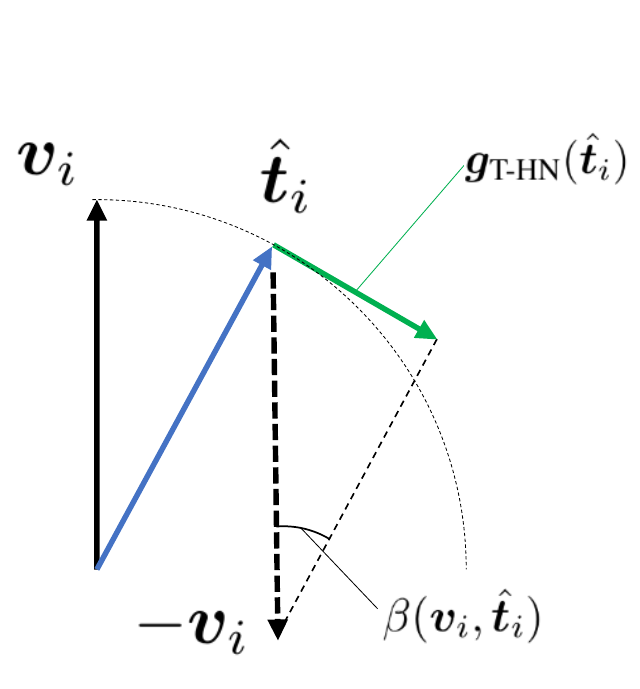}
		\caption{}
		\label{grad2}
	\end{subfigure}
	\hfill
	\begin{subfigure}{0.32\linewidth}
		\includegraphics[width=\linewidth]{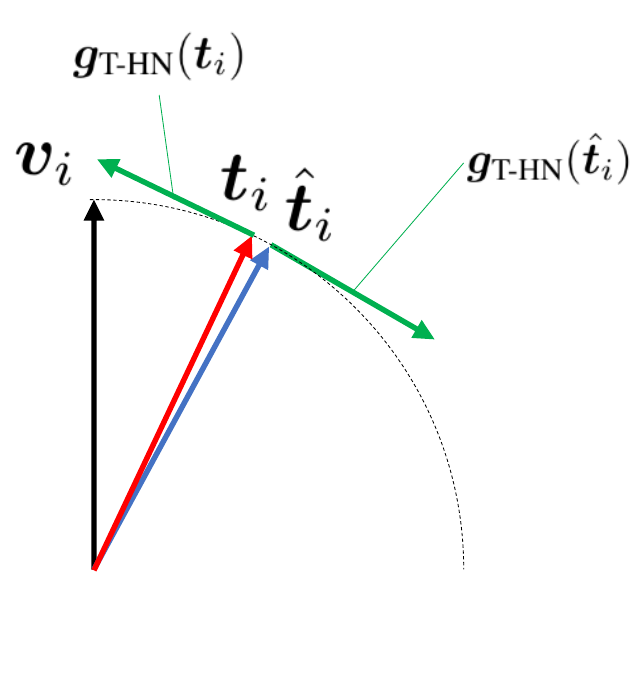}
		\caption{}
		\label{grad3}
	\end{subfigure}
	\vspace{-5pt}
	\caption{Schematic diagram of the gradient of the embeddings when the ITM model is trained with HN.}
	\vspace{-10pt}
	\label{gradient}
\end{figure}
\subsection{Gradient Vanishing in Image-Text Matching}
To explore why hard negative mining causes gradient vanishing, we analyze the gradient of  $ \mathcal{L}_{\text{T-HN}} $ on the learning of neural networks, and we get the condition under which the gradient vanishes during training.
We consider  $ \bm{f}_{vis}(\cdot) $ and $ \bm{f}_{text}(\cdot) $ embed the image and text on a unit hypersphere. 
To simplify the representation, we only analyze the loss in the image-to-text direction, which is symmetric with the loss in the text-to-image direction.
Taking $ \bm{v}_{i} $ as the anchor, the triplet loss with HN  takes the form of:
\begin{equation}
	\mathcal{L}_{\text{T-HN}}
	(\bm{v}_{i})
	= \max 
	\left( 
	\bm{v}_{i}^{\top} \hat{\bm{t}}_{i} 
	- \bm{v}_{i}^{\top} \bm{t}_{i} + \lambda, 0
	\right),
\end{equation}
where $ \bm{t}_{i} $ is the positive sample, $ \hat{\bm{t}}_{i} $ is the hard negative sample.
When $ \left( \bm{v}_{i}^{\top} \hat{\bm{t}}_{i} 
- \bm{v}_{i}^{\top} \bm{t}_{i} + \lambda \right) \leq 0 $, $ \mathcal{L}_{\text{T-HN}}
(\bm{v}_{i})=0 $.
Therefore, we only consider the case when $ \left( \bm{v}_{i}^{\top} \hat{\bm{t}}_{i} 
- \bm{v}_{i}^{\top} \bm{t}_{i} + \lambda \right) > 0 $.
We derive the loss gradient with respect to $ \bm{\theta}_{vis} $ and $ \bm{\theta}_{text} $:
\begin{equation}
	\frac{
		\partial 
		\mathcal{L}_{\text{T-HN}}
		(\bm{v}_{i})
	}
	{\partial \bm{\theta}_{vis}} 
	= 
	\frac{
		\partial 
		\mathcal{L}_{\text{T-HN}}
		(\bm{v}_{i})
	}
	{ \partial \bm{v}_{i} }
	\frac{ \partial \bm{v}_{i} }
	{ \partial \bm{\theta}_{vis} },
\end{equation}
\begin{equation}
	\frac{
		\partial 
		\mathcal{L}_{\text{T-HN}}
		(\bm{v}_{i})
	}
	{\partial \bm{\theta}_{text}} 
	= 
	\frac{
		\partial 
		\mathcal{L}_{\text{T-HN}}
		(\bm{v}_{i})
	}
	{ \partial \hat{\bm{t}}_{i} }
	\frac{ \partial \hat{\bm{t}}_{i} }
	{ \partial \bm{\theta}_{text} }
	+
	\frac{
		\partial 
		\mathcal{L}_{\text{T-HN}}
		(\bm{v}_{i})
	}
	{ \partial \bm{t}_{i} }
	\frac{ \partial \bm{t}_{i} }
	{ \partial \bm{\theta}_{text} }.
\end{equation}
The optimization of $ \bm{\theta}_{vis} $ is only related to the loss gradient with respect to $ \bm{v}_{i} $, and the optimization of $ \bm{\theta}_{text} $ is related to the loss gradient with respect to $ \hat{\bm{t}}_{i} $ and $ \bm{t}_{i} $.

The gradient optimization only affects the feature embedding through variations in $ \bm{\theta}_{vis} $ and $ \bm{\theta}_{text} $. 
But we first highlight problems with hypersphere embedding.
Therefore, we assume that the optimization could directly affect the embedding locations without considering the gradient effect caused by $ \bm{\theta}_{vis} $ and $ \bm{\theta}_{text} $.
We derive the loss gradient with respect to $ \bm{v}_{i} $, $ \hat{\bm{t}}_{i} $ and $ \bm{t}_{i} $:
\begin{equation}
	\frac{
		\partial 
		\mathcal{L}_{\text{T-HN}}
		(\bm{v}_{i})
	}
	{ \partial \bm{v}_{i} }
	= \hat{\bm{t}}_{i} - \bm{t}_{i},
	\frac{
		\partial 
		\mathcal{L}_{\text{T-HN}}
		(\bm{v}_{i})
	}
	{\partial \hat{\bm{t}}_{i}}
	= \bm{v}_{i},
	\frac{
		\partial 
		\mathcal{L}_{\text{T-HN}}
		(\bm{v}_{i})
	}
	{\partial \bm{t}_{i}} 
	= - \bm{v}_{i}.
\end{equation}
The loss gradient direction is related to the direction of $ \bm{v}_{i} $, $ \hat{\bm{t}}_{i} $ and $ \bm{t}_{i} $. 
The gradient can be decomposed into two components. 
One component of the hypersphere tangent direction contributes to the optimization of the model, such as the gradient component indicated by the green arrow in \figurename~\ref{grad2}.
The other is the component perpendicular to the tangent direction of the hypersphere.
This component pulls or pushes the features away from the hypersphere, and after $ L_{2} $ normalization it does not work for the model optimization.
We use $ \bm{g}_{\text{T-HN}}(\bm{v}_{i}) $, $ \bm{g}_{\text{T-HN}}(\hat{\bm{t}}_{i}) $ and $ \bm{g}_{\text{T-HN}}(\bm{t}_{i}) $ to denote the gradient components of $ \bm{v}_{i} $, $ \bm{t}_{i} $ and $ \hat{\bm{t}}_{i} $ that contribute to the model optimization, respectively.
Note that since the neural network is optimized using gradient descent, the directions of $ \bm{g}_{\text{T-HN}}(\bm{v}_{i}) $, $ \bm{g}_{\text{T-HN}}(\hat{\bm{t}}_{i}) $ and $ \bm{g}_{\text{T-HN}}(\bm{t}_{i}) $ are opposite to the direction of $ \frac{ \partial \mathcal{L}_{\text{T-HN}} (\bm{v}_{i}) }{ \partial \bm{v}_{i} } $, $ \frac{ \partial \mathcal{L}_{\text{T-HN}} (\bm{v}_{i}) }{ \partial \hat{\bm{t}}_{i} } $ and $ \frac{ \partial \mathcal{L}_{\text{T-HN}} (\bm{v}_{i}) }{ \partial \bm{t}_{i} } $.
We use $ \beta(\cdot, \cdot) $ to represent the angle between the two feature vectors, as shown in \figurename~\ref{grad1}.
The modulus of $ \bm{g}_{\text{T-HN}}(\bm{v}_{i}) $, $ \bm{g}_{\text{T-HN}}(\hat{\bm{t}}_{i}) $ and $ \bm{g}_{\text{T-HN}}(\bm{t}_{i}) $ can be expressed as:
\begin{equation}
	\begin{aligned}
	\left \Vert 
	\bm{g}_{\text{T-HN}}(\bm{v}_{i}) 
	\right \Vert 
	& =
	\left \Vert
	\frac{
		\partial 
		\mathcal{L}_{\text{T-HN}}
		(\bm{v}_{i})}
	{\partial \bm{v}_{i}} 
	\right \Vert
	\cdot 
	\sin 
	\beta(\bm{v}_{i}, \hat{\bm{t}}_{i} - \bm{t}_{i}) \\
	& =
	\left \Vert
	\hat{\bm{t}}_{i} - \bm{t}_{i}
	\right \Vert
	\cdot 
	\left( 
	\bm{v}_{i}^{\top} \hat{\bm{t}}_{i} 
	- \bm{v}_{i}^{\top} \bm{t}_{i}
	\right), \\
	\left \Vert 
	\bm{g}_{\text{T-HN}}(\hat{\bm{t}}_{i}) 
	\right \Vert 
	& =
	\left \Vert 
	\frac{
		\partial 
		\mathcal{L}_{\text{T-HN}}
		(\bm{v}_{i})}
	{\partial \hat{\bm{t}}_{i}} 
	\right \Vert 
	\cdot 
	\sin 
	\beta(\bm{v}_{i}, \hat{\bm{t}}_{i}) \\
	& = 
	\left \Vert 
	\bm{v}_{i}
	\right \Vert 
	\cdot
	\bm{v}_{i}^{\top} \hat{\bm{t}}_{i}
	= \bm{v}_{i}^{\top} \hat{\bm{t}}_{i}, \\
	\left \Vert 
	\bm{g}_{\text{T-HN}}(\bm{t}_{i}) 
	\right \Vert 
	& =
	\left \Vert 
	\frac{
		\partial 
		\mathcal{L}_{\text{T-HN}}
		(\bm{v}_{i})}
	{\partial \bm{t}_{i}} 
	\right \Vert 
	\cdot 
	\sin 
	\beta(\bm{v}_{i}, \bm{t}_{i}) \\
	& = 
	\left \Vert 
	- \bm{v}_{i}
	\right \Vert 
	\cdot
	\bm{v}_{i}^{\top} \bm{t}_{i}
	= \bm{v}_{i}^{\top} \bm{t}_{i}.
	\end{aligned}
\end{equation}
We observe that modulus of $ \bm{g}_{\text{T-HN}}(\bm{v}_{i}) $, $ \bm{g}_{\text{T-HN}}(\hat{\bm{t}}_{i}) $ and $ \bm{g}_{\text{T-HN}}(\bm{t}_{i}) $ are closely related to the similarity of positive pairs $ \bm{v}_{i}^{\top} \bm{t}_{i} $ and the similarity of negative pairs $ \bm{v}_{i}^{\top} \hat{\bm{t}}_{i} $.

\begin{figure}
	\centering
	\begin{subfigure}{0.49\linewidth}
		\includegraphics[width=\linewidth]{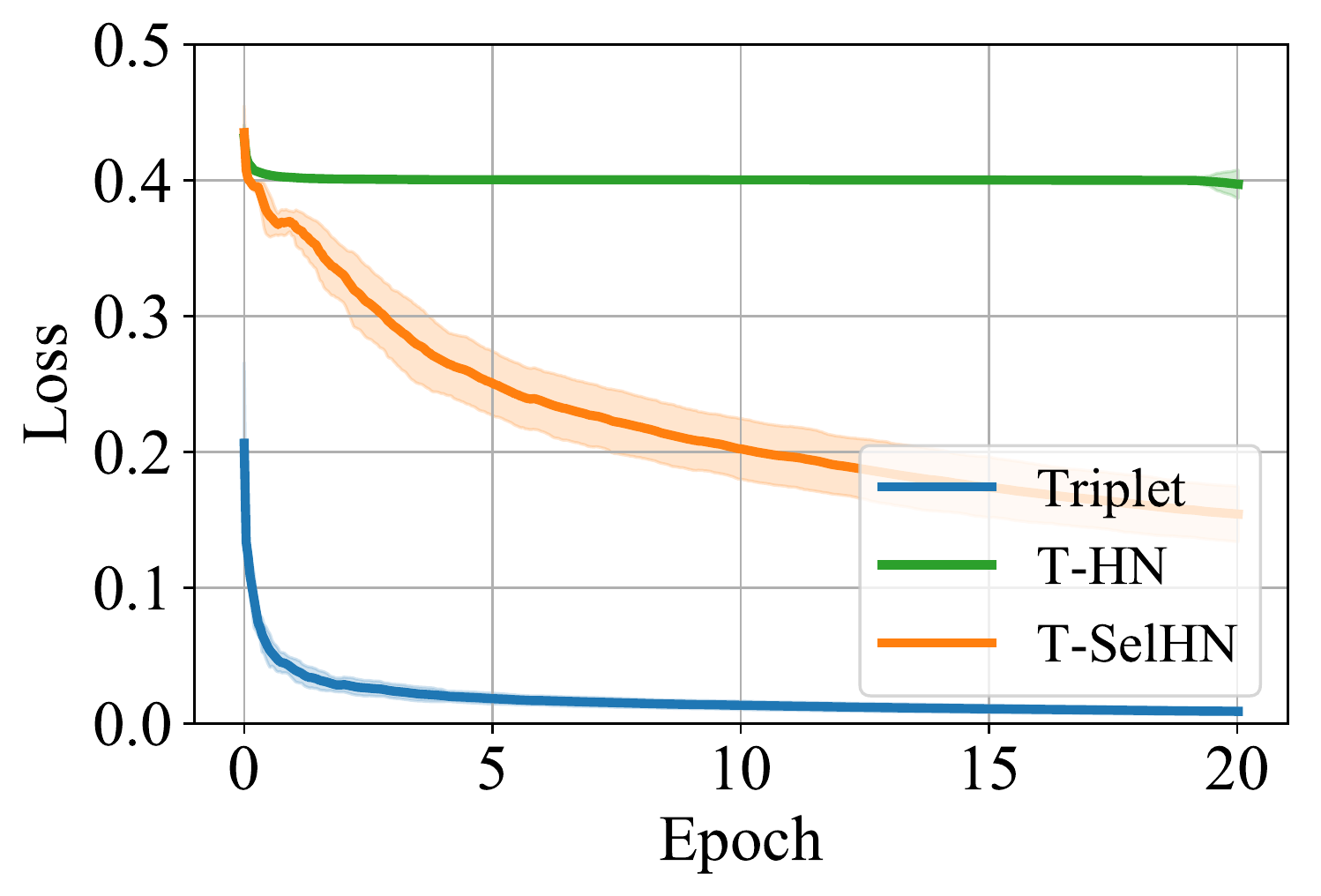}
		\caption{Loss}
		\label{train_loss_mlp}
	\end{subfigure}
	\hfill
	\begin{subfigure}{0.49\linewidth}
		\includegraphics[width=\linewidth]{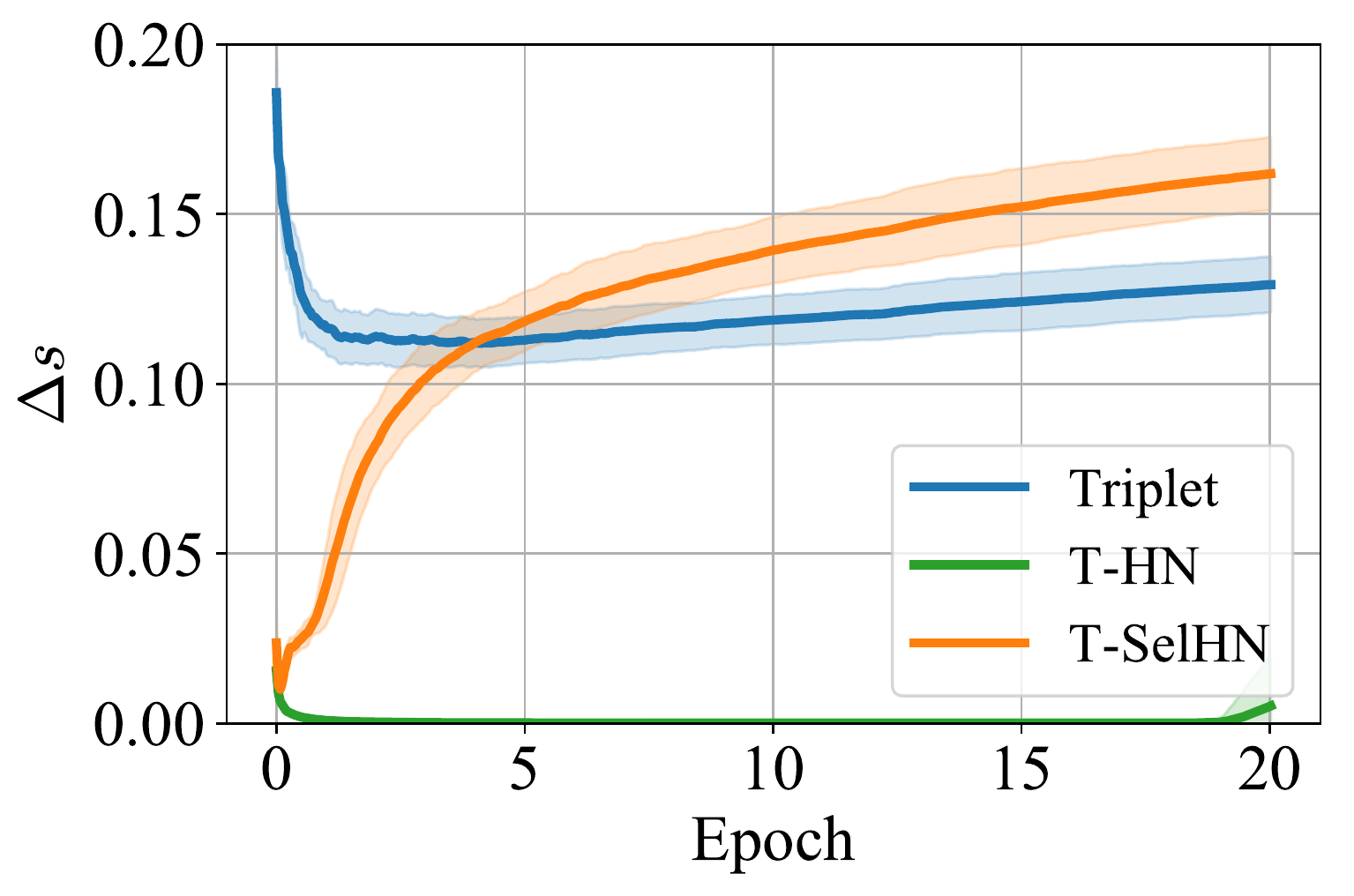}
		\caption{$ \Delta s $}
		\label{train_diff_mlp}
	\end{subfigure}
	\vspace{-5pt}
	\caption{Training performance of VSE (MLP) on Flickr30K.}
	\vspace{-10pt}
	\label{train_mlp}
\end{figure}
We define $ \Delta s = \left| \bm{v}_{i}^{\top} \hat{\bm{t}}_{i} - \bm{v}_{i}^{\top} \bm{t}_{i} \right| $.
We count $ \Delta s $ during training of the VSE (FC), as shown in \figurename~\ref{train_diff_fc}.
At the beginning of training (epoch 0-7), $ \Delta s $ approaches $ 0 $.
When $ \Delta s \rightarrow 0 $, $ \left \Vert \bm{g}_{\text{T-HN}}(\bm{v}_{i}) \right \Vert \rightarrow 0 $.
The optimization of $ \bm{\theta}_{vis} $ is only related to the gradient of $ \bm{v}_{i} $. 
Therefore, there is a gradient vanishing in the optimization of the image encoder.
On the other hand, the optimization of $ \bm{\theta}_{text} $ is related to $ \hat{\bm{t}}_{i} $ and $ \bm{t}_{i} $. 
When $ \Delta s \rightarrow 0 $, $ \bm{g}_{\text{T-HN}}(\hat{\bm{t}}_{i}) $ and $ \bm{g}_{\text{T-HN}}(\bm{t}_{i}) $ are equal in magnitude and opposite in direction, as shown in \figurename~\ref{grad3}. 
These two gradients cancel each other out when optimizing $ \bm{\theta}_{text} $.
Therefore, there is also a gradient vanishing in the optimization of the text encoder.
Our gradient analysis and experimental results yield a consistent condition under which gradient vanishing occurs.
When $ \Delta s \rightarrow 0 $, the training of the ITM model suffers from the gradient vanishing.

\begin{table}[t] 
	\setlength\tabcolsep{2pt}
	\footnotesize
	\begin{center}
		\begin{tabular}{llccccccccccc}
			\toprule[1pt]
			\multicolumn{2}{l}{Eval Task $ \rightarrow $}
			& \multicolumn{3}{c}{Image-to-Text} & \multicolumn{3}{c}{Text-to-Image} & \multirow{2}*{RSUM}\\
			\cline{1-8}
			\specialrule{0em}{2pt}{0pt}
			Model $ \downarrow $ & Loss $ \downarrow $ & R@1 & R@5 & R@10 & R@1 & R@5 & R@10 \\
			\hline
			
			\specialrule{0em}{2pt}{0pt}
			VSE (FC) & $ \mathcal{L}_{\text{Triplet}} $ & 64.5 & 89.5 & 95.0 & 48.4 & 77.4 & 86.0 & 460.7 \\
			VSE (FC) & $ \mathcal{L}_{\text{T-HN}} $ & 76.0 & 92.5 & 97.0 & 53.9 & 81.2 & 88.2 & 488.8 \\
			\rowcolor{black!10}
			VSE (FC) & $ \mathcal{L}_{\text{T-SelHN}} $ & 77.1 & 93.6 & 97.1 & 56.3 & 82.8 & 89.2 & \textbf{496.1} \\
			\hline
			
			\specialrule{0em}{2pt}{0pt}
			VSE (MLP) & $ \mathcal{L}_{\text{Triplet}} $ & 59.1 & 86.0 & 92.8 & 48.4 & 77.2 & 85.9 & 449.4 \\
			VSE (MLP) & $ \mathcal{L}_{\text{T-HN}} $ & 44.0 & 70.0 & 79.1 & 31.9 & 61.6 & 72.9 & 359.4 \\
			\rowcolor{black!10}
			VSE (MLP) & $ \mathcal{L}_{\text{T-SelHN}} $ & 75.5 & 92.8 & 96.9 & 55.9 & 82.4 & 89.3 & \textbf{492.8} \\
			\hline
			
			\specialrule{0em}{2pt}{0pt}
			RVSE (MLP) & $ \mathcal{L}_{\text{Triplet}} $ & 64.8 & 89.3 & 95.0 & 49.9 & 78.5 & 87.1 & 464.6 \\
			RVSE (MLP) & $ \mathcal{L}_{\text{T-HN}} $ & 72.4 & 92.7 & 96.8 & 54.1 & 80.9 & 87.7 & 484.6 \\
			\rowcolor{black!10}
			RVSE (MLP) & $ \mathcal{L}_{\text{T-SelHN}} $ & 77.4 & 94.6 & 97.3 & 56.8 & 82.8 & 89.7 & \textbf{498.6} \\
			\bottomrule[1pt]
		\end{tabular}
	\end{center}
	\vspace{-15pt}
	\caption{Matching performance of three ITM models trained with different loss functions on Flickr30K test set.}
	\vspace{-10pt}
	\label{f30k_bigru}
\end{table}
The experimental results on other models also show a consistent conclusion.
The training performance of the VSE (MLP) model is shown in \figurename~\ref{rvse_mlp}. 
Using $ \mathcal{L}_{\text{T-HN}} $ for training suffers from serious gradient vanishing, $ \Delta s $ always approaches 0, and the $ \mathcal{L}_{\text{T-HN}} $ decreases very slowly.
Since the VSE (MLP) model has more layers than VSE (FC), it has been shown that with the increase of the number of layers, the gradient vanishing is prone to occur when using the gradient descent method to optimize neural networks~\cite{he2016deep}.
\tablename~\ref{f30k_bigru} shows the matching performance of several models trained with different loss functions. 
Due to the severe gradient vanishing, the matching performance of VSE (MLP) trained with $ \mathcal{L}_{\text{T-HN}} $ suffers a large drop. 
This shows that the gradient vanishing can make the model unable to fully exert its representation power, which can negatively affect the matching performance of the model.

\subsection{Selectively Hard Negative Mining}
$ \mathcal{L}_{\text{Triplet}} $ and $ \mathcal{L}_{\text{T-HN}} $ have their own pros and cons. 
$ \mathcal{L}_{\text{T-HN}} $ can achieve better matching performance than $ \mathcal{L}_{\text{Triplet}} $ by hard negative mining. 
Compared with $ \mathcal{L}_{\text{T-HN}} $, $ \mathcal{L}_{\text{Triplet}} $ does not suffer from gradient vanishing during training.
To take full advantage of hard negative mining and alleviate the gradient vanishing in model training, we propose a Selectively Hard Negative Mining (SelHN) strategy.
SelHN chooses whether to mine hard negative samples according to the gradient vanishing condition.
With $ V_{i} $ as the anchor, the triplet loss with SelHN takes the form of:
\begin{equation}
	\mathcal{L}_{\text{T-SelHN}}
	(V_{i}) 
	=
	\begin{cases}
		\mathcal{L}_{\text{T-HN}}
		(V_{i}) ,
		& \Delta s(V_{i})
		> \epsilon, \\
		\mathcal{L}_{\text{Triplet}}
		(V_{i}) , 
		& \text{otherwise},
	\end{cases}
\end{equation}
where
\begin{equation}
	\Delta s(V_{i})
	= \left| s(V_{i}, \hat{T}_{i}) - s(V_{i}, T_{i}) \right|,
\end{equation}
\begin{equation}
	\mathcal{L}_{\text{T-HN}}(V_{i})
	=
	[ 
	s(V_{i}, \hat{T}_{i}) - s(V_{i}, T_{i}) + \lambda 
	]_{+},
\end{equation}
\begin{equation}
	\mathcal{L}_{\text{Triplet}}(V_{i})
	=
	\dfrac{1}{B}
	\sum_{j=1, i \neq j}^{B}
	[ 
	s(V_{i}, T_{j}) - s(V_{i}, T_{i}) + \lambda 
	]_{+},
\end{equation}
$ \epsilon $ is a threshold used to judge whether the loss should mine hard negative samples.
$ \epsilon $ takes the value of a small positive number. 
In our experiments, $ \epsilon $ is taken as 0.01.
When $ \Delta s(V_{i}) > \epsilon $, the model training is not easy to suffer from the gradient vanishing, and the discriminative power of the model can be improved by hard negative mining. 
When $ \Delta s(V_{i}) < \epsilon $, the model is optimized using $ \mathcal{L}_{\text{Triplet}}(V_{i}) $ to increase $ \Delta s(V_{i}) $.
As shown in \figurename~\ref{fc} and \figurename~\ref{train_mlp}, $ \mathcal{L}_{\text{T-SelHN}} $ exhibits excellent training performance. 
Using $ \mathcal{L}_{\text{T-SelHN}} $ as the optimization objective, the training of the model does not suffer from gradient vanishing, and can achieve the best matching performance faster than $ \mathcal{L}_{\text{Triplet}} $ and $ \mathcal{L}_{\text{T-HN}} $.

\subsection{Residual Visual Semantic Embedding}
SelHN can alleviate the gradient vanishing in model training and take advantage of hard negative sample mining. 
\tablename~\ref{f30k_bigru} shows that the matching performance of $ \mathcal{L}_{\text{T-SelHN}} $ is significantly better than $ \mathcal{L}_{\text{Triplet}} $ and $ \mathcal{L}_{\text{T-HN}} $ on VSE (FC) and VSE (MLP). 
However, we observe that the performance on VSE (MLP) with more layers is overall inferior to the simpler VSE (FC) model. 
This is because as the number of layers in the model increases, gradient vanishing is more likely to occur in the back-propagation process of the gradient.
Residual connection is an effective way to alleviate the gradient vanishing. 
\tablename~\ref{f30k_bigru} shows that RVSE (MLP) training with $ \mathcal{L}_{\text{T-SelHN}} $ can achieve the best performance.

To achieve better matching performance, we construct a Residual Visual Semantic Embedding model with SelHN, denote as RVSE++.
The image encoder of RVSE++ uses the RVSE (MLP) structure shown in \figurename~\ref{rvse_mlp}. 
The text encoder uses a more representative BERT model~\cite{kenton2019bert} to extract word-level features $ \{ \bm{w}_{i}^{n} \}_{n=1}^{N} $.
Then we use a FC layer to map the dimension of $ \{ \bm{w}_{i}^{n} \}_{n=1}^{N} $ to $ d $ and use a GPO to aggregate the text feature set into the text embedding $ \bm{t}_{i} $.
Finally, we use cosine similarity to calculate the match score between the image and text.
RVSE++ has a simple network structure, few parameters, and fast inference speed. 
Thanks to our network design and SelHN strategy, the representation ability of RVSE++ can be fully exerted.
Subsequent experiments verify the strength of RVSE++ in many aspects.

\section{Experiments}
\subsection{Dataset and Experiment Settings}
\textbf{Datasets.}
We evaluate our SelHN and RVSE++ on two benchmarks: Flickr30K~\cite{young2014image} and MS-COCO~\cite{lin2014microsoft}.
Flickr30K contains 31,000 images, each image is annotated with 5 sentences.
We use 1,000 images for validation, 1,000 images for testing and the remaining 29,000 images for training. 
MS-COCO contains 123,287 images, each image associates with 5 sentences. 
We use 113,287 images for training, 5,000 images for validation and 5,000 images for testing. 
We report results on both 1,000 test images (averaged over 5 folds) and full 5,000 test images of MS-COCO.

\textbf{Evaluation Metrics.}
For the evaluation on Flickr30K and MS-COCO, following the~\cite{faghri2018vse++}, we use the Recall@K (R@K), with $ K = \{1,5,10\} $ as the evaluation metric for the task. 
R@K indicates the percentage of queries for which the model returns the correct item in its top $ K $ results. 
We follow~\cite{chen2021learning} to use RSUM, which is defined as the sum of recall metrics at $ K = \{1,5,10\} $ of both text-to-image and image-to-text matching, as an average metric of the overall performance of the ITM model.

\textbf{Implementation Details.}
Our all experiments are conducted on an NVIDIA GeForce RTX 3090 GPU using PyTorch.
For SelHN, hyper-parameters are set as $ \epsilon = 0.01 $ and $ \lambda = 0.2 $  for both models and datasets. 
Our RVSE++ is trained using AdamW for 30 epochs, with a batch size of 128 for both datasets. 
The initial learning rate of the model is set as 0.0005 for the first 15 epochs and then decays by a factor of 10 for the last 15 epochs. 
More implementation details and experimental results are listed in the appendix.

\subsection{Improvements on Existing ITM Models}
\begin{table}[t]
	\setlength\tabcolsep{2pt}
	\footnotesize
	\begin{center}
		\begin{tabular}{lcccccccccccc}
			\toprule[1pt]
			Eval Task $ \rightarrow $
			& \multicolumn{3}{c}{Image-to-Text} & \multicolumn{3}{c}{Text-to-Image} & \multirow{2}*{RSUM}\\
			\cline{1-7}
			\specialrule{0em}{2pt}{0pt}
			Method $ \downarrow $ & R@1 & R@5 & R@10 & R@1 & R@5 & R@10 \\
			\hline
			
			\specialrule{0em}{2pt}{0pt}
			SCAN$ _{\text{(\textit{ECCV}'18)}}$~\cite{lee2018stacked} & 67.4 & 90.3 & 95.8 & 48.6 & 77.7 & 85.2 & 465.0 \\
			\rowcolor{black!10}
			+ SelHN & 73.4 & 93.7 & 96.6 & 53.5 & 81.1 & 87.9 & \textbf{486.2} \\
			\hline

			\specialrule{0em}{2pt}{0pt}
			BFAN$ _{\text{(\textit{MM}'19)}}$~\cite{liu2019focus} & 68.1 & 91.4 & 95.9 & 50.8 & 78.4 & 85.8 & 470.4 \\
			\rowcolor{black!10}
			+ SelHN & 75.3 & 93.4 & 97.1 & 55.2 & 80.9 & 87.3 & \textbf{489.1} \\
			\hline

			\specialrule{0em}{2pt}{0pt}
			SGRAF$ _{\text{(\textit{AAAI}'21)}}$~\cite{diao2021similarity} & 77.8 & 94.1 & 97.4 & 58.5 & 83.0 & 88.8 & 499.6 \\
			\rowcolor{black!10}
			+ SelHN & 80.0 & 95.1 & 98.2 & 59.8 & 84.5 & 89.5 & \textbf{507.1} \\
			
			\bottomrule[1pt]
		\end{tabular}
	\end{center}
	\vspace{-15pt}
	\caption{Matching performance of applying SelHN to existing ITM models on Flickr30K test set.}
	\vspace{-10pt}
	\label{exist}
\end{table}
\begin{table*}[t] 
	\setlength\tabcolsep{4pt}
	\footnotesize
	\begin{center}
		\begin{tabular}{lcccccccccccccccccccc}
			\toprule[1pt]
			Eval Task $ \rightarrow $
			& \multicolumn{3}{c}{Image-to-Text} & \multicolumn{3}{c}{Text-to-Image} & \multirow{2}*{RSUM} & & \multicolumn{3}{c}{Image-to-Text} & \multicolumn{3}{c}{Text-to-Image} & \multirow{2}*{RSUM} \\
			\cline{1-7}\cline{10-15}
			\specialrule{0em}{2pt}{0pt}
			Loss $ \downarrow $ & R@1 & R@5 & R@10 & R@1 & R@5 & R@10 & & ~ & R@1 & R@5 & R@10 & R@1 & R@5 & R@10 \\
			\hline
			
			\specialrule{0em}{2pt}{0pt}
			$ \mathcal{L}_{\text{Triplet}} $ & 64.5 & 89.5 & 95.0 & 48.4 & 77.4 & 86.0 & 460.7 & & 41.4 & 72.5 & 83.3 & 30.6 & 60.0 & 73.0 & 360.7 \\
			$ \mathcal{L}_{\text{T-HN}} $ & 76.0 & 92.5 & 97.0 & 53.9 & 81.2 & 88.2 & 488.8 & & 54.3 & 82.3 & 90.2 & 37.3 & 67.6 & 79.0 & 410.7 \\
			SHN & 74.0 & 93.4 & 97.4 & 55.3 & 82.2 & 88.8 & 491.0 & & 50.4 & 79.8 & 88.7 & 35.8 & 66.1 & 78.2 & 399.1 \\
			SCT & 75.8 & 93.6 & 97.1 & 55.3 & 82.0 & 88.6 & 492.3 & & 54.6 & 82.4 & 90.2 & 36.5 & 67.1 & 78.9 & 409.7 \\
			\rowcolor{black!10}
			$ \mathcal{L}_{\text{T-SelHN}} $ & 77.1 & 93.6 & 97.1 & 56.3 & 82.8 & 89.2 & \textbf{496.1} & & 54.6 & 82.7 & 90.5 & 38.0 & 68.5 & 80.1 & \textbf{414.4} \\
			
			\bottomrule[1pt]
		\end{tabular}
	\end{center}
	\vspace{-15pt}
	\caption{Performance comparison with other methods that improve training behavior on Flickr30K test set and MS-COCO 5K test set.}
	\vspace{-5pt}
	\label{other_triplet}
\end{table*}
To justify the superiority of our SelHN over the existing ITM models, we conduct experiments on SCAN~\cite{lee2018stacked}, BFAN~\cite{liu2019focus}, and SGRAF~\cite{diao2021similarity} by only replacing the loss functions.
SCAN and BFAN belong to the CA method, and SGRAF belongs to the combined method of VSE and CA.
Our previous experiments and gradient analysis are mainly based on the VSE model. 
This experiment shows that the conclusions we obtain can be extended to other types of models.
\tablename~\ref{exist} shows the improvements on these models on the Flickr30K dataset.
By replacing the loss function with our SelHN, the performance of the three baseline methods is improved.
On SCAN and BFAN, the models using SelHN are significantly improved by 21.2\% and 18.7\% on RSUM, respectively, compared with the original models.
On the more advanced SGRAF, the RSUM is also improved by 7.5\% after applying SelHN.
Our SelHN strategy can be plug-and-play applied to existing ITM models. 
The experimental results show that the application of SelHN can give full play to the representation ability of the model and achieve better matching performance.

\subsection{Comparisons with the Other Loss Functions}
Our SelHN focuses on alleviating gradient vanishing in ITM model training. 
Therefore, we compare SelHN with other methods that improve training behavior:
\begin{itemize}[leftmargin=*]
	\item \textbf{SHN}: Semi-Hard Negative mining~\cite{schroff2015facenet} does not optimize hard negative samples, and only mines semi-hard negative samples with $ s_{n}<s_{p} $ for optimization.
	\item \textbf{SCT}: Selectively Contrastive Triplet loss~\cite{xuan2020hard} uses the contrastive loss to optimize hard negative samples and uses the triplet loss to optimize the remaining samples.
\end{itemize}
The experimental results of SelHN compared to the above losses on the VSE (FC) model are shown in \tablename~\ref{other_triplet}.
Compared with these loss functions, SelHN improves most evaluation metrics.
SHN and SCT improve the training behavior. 
But they abandon optimization for triplet consisting of hard negative samples.
Hard negative samples are crucial for learning a discriminative model. 
SelHN mines hard negative samples when gradient vanishing is unlikely to occur. 
Therefore, SelHN not only improves the training behavior but also gives full play to the role of hard negative mining.

\begin{table}[t] 
	\setlength\tabcolsep{2pt}
	\footnotesize
	\begin{center}
		\begin{tabular}{llccccccccccc}
			\toprule[1pt]
			\multicolumn{2}{l}{Eval Task $ \rightarrow $}
			& \multicolumn{3}{c}{Image-to-Text} & \multicolumn{3}{c}{Text-to-Image} & \multirow{2}*{RSUM}\\
			\cline{1-8}
			\specialrule{0em}{2pt}{0pt}
			Model $ \downarrow $ & Loss $ \downarrow $ & R@1 & R@5 & R@10 & R@1 & R@5 & R@10 \\
			\hline

			\specialrule{0em}{2pt}{0pt}
			BFAN & $ \mathcal{L}_{\text{T-HN}} $ & 68.1 & 91.4 & 95.9 & 50.8 & 78.4 & 85.8 & 470.4 \\
			BFAN & SSP & 71.3 & 92.6 & 96.2 & 52.5 & 79.5 & 86.6 & 478.7 \\
			BFAN & AOQ & 73.2 & 94.5 & 97.0 & 54.0 & 80.3 & 87.7 & 486.7 \\
			BFAN & Meta-SPN & 72.5 & 93.2 & 96.7 & 53.3 & 80.2 & 87.2 & 483.1 \\
			\rowcolor{black!10}
			BFAN & $ \mathcal{L}_{\text{T-SelHN}} $ & 75.3 & 93.4 & 97.1 & 55.2 & 80.9 & 87.3 & \textbf{489.1} \\
			\hline
			
			\specialrule{0em}{2pt}{0pt}
			SGRAF & $ \mathcal{L}_{\text{T-HN}} $ & 77.8 & 94.1 & 97.4 & 58.5 & 83.0 & 88.8 & 499.6 \\
			SGRAF & NCR & 77.3 & 94.0 & 97.5 & 59.6 & 84.4 & 89.9 & 502.7 \\
			\rowcolor{black!10}
			SGRAF & $ \mathcal{L}_{\text{T-SelHN}} $ & 80.0 & 95.1 & 98.2 & 59.8 & 84.5 & 89.5 & \textbf{507.1} \\
			
			\bottomrule[1pt]
		\end{tabular}
	\end{center}
	\vspace{-15pt}
	\caption{Performance comparison with other loss functions proposed for ITM on Flickr30K test set.}
	\vspace{-10pt}
	\label{other_itm}
\end{table}
On the other hand, there are several loss functions proposed for ITM, so we compare SelHN with these losses:
\begin{itemize}[leftmargin=*]
	\item \textbf{SSP}: Self-Similarity Polynomial loss~\cite{wei2020universal} is a weighted triplet loss that defines a weight function for the positive and negative pairs respectively.
	\item \textbf{Meta-SPN}: Meta Self-Paced Network~\cite{wei2021meta} automatically learns a weighting scheme for ITM.
	\item \textbf{AOQ}: Adaptive Offline Quintuplet loss \cite{chen2020adaptive} mines offline negatives from the whole training set.
	\item \textbf{NCR}: Noisy Correspondence Rectifier \cite{huang2021learning} focuses on learning with noisy correspondence for ITM.
\end{itemize}
For a fair comparison, we use the same model as in the papers of various loss functions above but replace the loss function with our SelHN.
The experimental results of SelHN compared to the other losses on Flickr30K are shown in \tablename~\ref{other_itm}.
Compared with other loss functions, SelHN improves most evaluation metrics.
SelHN does not need to introduce many hyperparameters to assign weights like SSP, nor does it need to train an additional network for weight assignment like Meta-SPN.
Compared with AOQ, SelHN only requires online hard negative mining and does not increase the complexity.

\subsection{Comparisons with state-of-the-art Methods}
\begin{table}[t] 
	\setlength\tabcolsep{2pt}
	\footnotesize
	\begin{center}
		\begin{tabular}{lcccccccccccc}
			\toprule[1pt]
			Eval Task $ \rightarrow $
			& \multicolumn{3}{c}{Image-to-Text} & \multicolumn{3}{c}{Text-to-Image} & \multirow{2}*{RSUM}\\
			\cline{1-7}
			\specialrule{0em}{2pt}{0pt}
			Method $ \downarrow $ & R@1 & R@5 & R@10 & R@1 & R@5 & R@10 \\
			\hline
			\specialrule{0em}{2pt}{0pt}
			SCAN$ _{\text{(\textit{ECCV}'18)}} $~\cite{lee2018stacked} & 67.4 & 90.3 & 95.8 & 48.6 & 77.7 & 85.2 & 465.0 \\
			CAMP$ _{\text{(\textit{CVPR}'19)}} $~\cite{wang2019camp} & 68.1 & 89.7 & 95.2 & 51.5 & 77.1 & 85.3 & 466.9 \\
			BFAN$ _{\text{(\textit{MM}'19)}} $~\cite{liu2019focus} & 68.1 & 91.4 & 95.9 & 50.8 & 78.4 & 85.8 & 470.4 \\
			VSRN$ _{\text{(\textit{ICCV}'19)}} $~\cite{li2019visual} & 71.3 & 90.6 & 96.0 & 54.7 & 81.8 & 88.2 & 482.6 \\
			CVSE$ _{\text{(\textit{ECCV}'20)}} $~\cite{wang2020consensus} & 73.5 & 92.1 & 95.8 & 52.9 & 80.4 & 87.8 & 482.5 \\
			IMRAM$ _{\text{(\textit{CVPR}'20)}} $~\cite{chen2020imram} &  74.1 & 93.0 & 96.6 & 53.9 & 79.4 & 87.2 & 484.2 \\
			GSMN$ _{\text{(\textit{CVPR}'20)}} $~\cite{liu2020graph} & 76.4 & 94.3 & 97.3 & 57.4 & 82.3 & 89.0 & 496.8 \\
			SGRAF$ _{\text{(\textit{AAAI}'21)}} $~\cite{diao2021similarity} & 77.8 & 94.1 & 97.4 & 58.5 & 83.0 & 88.8 & 499.6 \\
			VSE$\infty$ $ _{\text{(\textit{CVPR}'21)}} $~\cite{chen2021learning} & 81.7 & 95.4 & 97.6 & 61.4 & 85.9 & 91.5 & 513.5 \\
			CMCAN$ _{\text{(\textit{AAAI}'22)}} $~\cite{zhang2022show} & 79.5 & 95.6 & 97.6 & 60.9 & 84.3 & 89.9 & 507.8 \\
			NAAF$ _{\text{(\textit{CVPR}'22)}} $~\cite{zhang2022negative} & 81.9 & 96.1 & 98.3 & 61.0 & 85.3 & 90.6 & 513.2 \\
			\hline
			
			\specialrule{0em}{2pt}{0pt}
			\rowcolor{black!10}
			\textbf{RVSE++} & 83.6 & 96.5 & 98.6 & 64.3 & 88.2 & 93.0 & \textbf{524.2} \\
			\bottomrule[1pt]
		\end{tabular}
	\end{center}
	\vspace{-15pt}
	\caption{Performance comparison with the state-of-the-art ITM methods on Flickr30K test set.}
	\vspace{-10pt}
	\label{f30k}
\end{table}
We compare our proposed RVSE++ with the state-of-the-art methods on the two benchmarks.
For a fair comparison, the methods we compare all use the BUTD features~\cite{anderson2018bottom}.
\tablename~\ref{f30k} shows the quantitative results of RVSE++ on Flickr30K. 
Our RVSE++ outperforms these methods significantly on all evaluation metrics. 
Specifically, compared with the recent state-of-the-art method NAAF, RVSE++ obtain a relative 11\% improvement on RSUM.

\begin{table}[t] 
	\setlength\tabcolsep{2pt}
	\footnotesize
	\begin{center}
		\begin{tabular}{lcccccccccccc}
			\toprule[1pt]
			Eval Task $ \rightarrow $
			& \multicolumn{3}{c}{Image-to-Text} & \multicolumn{3}{c}{Text-to-Image} & \multirow{2}*{RSUM}\\
			\cline{1-7}
			\specialrule{0em}{2pt}{0pt}
			Method $ \downarrow $ & R@1 & R@5 & R@10 & R@1 & R@5 & R@10 \\
			\hline
			\specialrule{0em}{2pt}{0pt}
			SCAN$ _{\text{(\textit{ECCV}'18)}} $~\cite{lee2018stacked} & 72.7 & 94.8 & 98.4 & 58.8 & 88.4 & 94.8 & 507.9 \\
			CAMP$ _{\text{(\textit{CVPR}'19)}} $~\cite{wang2019camp} & 72.3 & 94.8 & 98.3 & 58.5 & 87.9 & 95.0 & 506.8 \\ 
			BFAN$ _{\text{(\textit{MM}'19)}} $~\cite{liu2019focus} & 74.9 & 95.2 & 98.3 & 59.4 & 88.4 & 94.5 & 510.7 \\
			VSRN$ _{\text{(\textit{ICCV}'19)}} $~\cite{li2019visual} & 76.2 & 94.8 & 98.2 & 62.8 & 89.7 & 95.1 & 516.8 \\
			CVSE$ _{\text{(\textit{ECCV}'20)}} $~\cite{wang2020consensus} & 74.8 & 95.1 & 98.3 & 59.9 & 89.4 & 95.2 & 512.7 \\
			IMRAM$ _{\text{(\textit{CVPR}'20)}} $~\cite{chen2020imram} & 76.7 & 95.6 & 98.5 & 61.7 & 89.1 & 95.0 & 516.6 \\
			GSMN$ _{\text{(\textit{CVPR}'20)}} $~\cite{liu2020graph} & 78.4 & 96.4 & 98.6 & 63.3 & 90.1 & 95.7 & 522.5 \\
			SGRAF$ _{\text{(\textit{AAAI}'21)}} $~\cite{diao2021similarity} & 79.6 & 96.2 & 98.5 & 63.2 & 90.7 & 96.1 & 524.3 \\
			VSE$\infty$ $ _{\text{(\textit{CVPR}'21)}} $~\cite{chen2021learning} & 79.7 & 96.4 & 98.9 & 64.8 & 91.4 & 96.3 & 527.5 \\
			CMCAN$ _{\text{(\textit{AAAI}'22)}} $~\cite{zhang2022show} & 81.2 & 96.8 & 98.7 & 65.4 & 91.0 & 96.2 & 529.3 \\
			NAAF$ _{\text{(\textit{CVPR}'22)}} $~\cite{zhang2022negative} & 80.5 & 96.5 & 98.8 & 64.1 & 90.7 & 96.5 & 527.2 \\
			\hline
			
			\specialrule{0em}{2pt}{0pt}
			\rowcolor{black!10}
			\textbf{RVSE++} & 81.6 & 96.6 & 98.8 & 66.6 & 92.1 & 96.6 & \textbf{532.4} \\
			\bottomrule[1pt]
		\end{tabular}
	\end{center}
	\vspace{-15pt}
	\caption{Performance comparison with the state-of-the-art ITM methods on MS-COCO 1K test set.}
	\label{coco1k}
\end{table}
The experimental results on the MS-COCO 1K test set are shown in \tablename~\ref{coco1k}. 
We can see that our RVSE++ outperforms these methods in terms of most evaluation metrics. 
Compared with NAAF, RVSE++ gains relative improvements of 5.2\% on RSUM. 
Experimental results demonstrate the effectiveness of our RVSE++. 
The residual design of the RVSE++ model and the guarantee of the SelHN strategy for gradient backpropagation enable the RVSE++ model to give full play to the representation ability of the model.

\subsection{Ablation Study}
\begin{figure}
	\centering
	\begin{subfigure}{0.49\linewidth}
		\includegraphics[width=\linewidth]{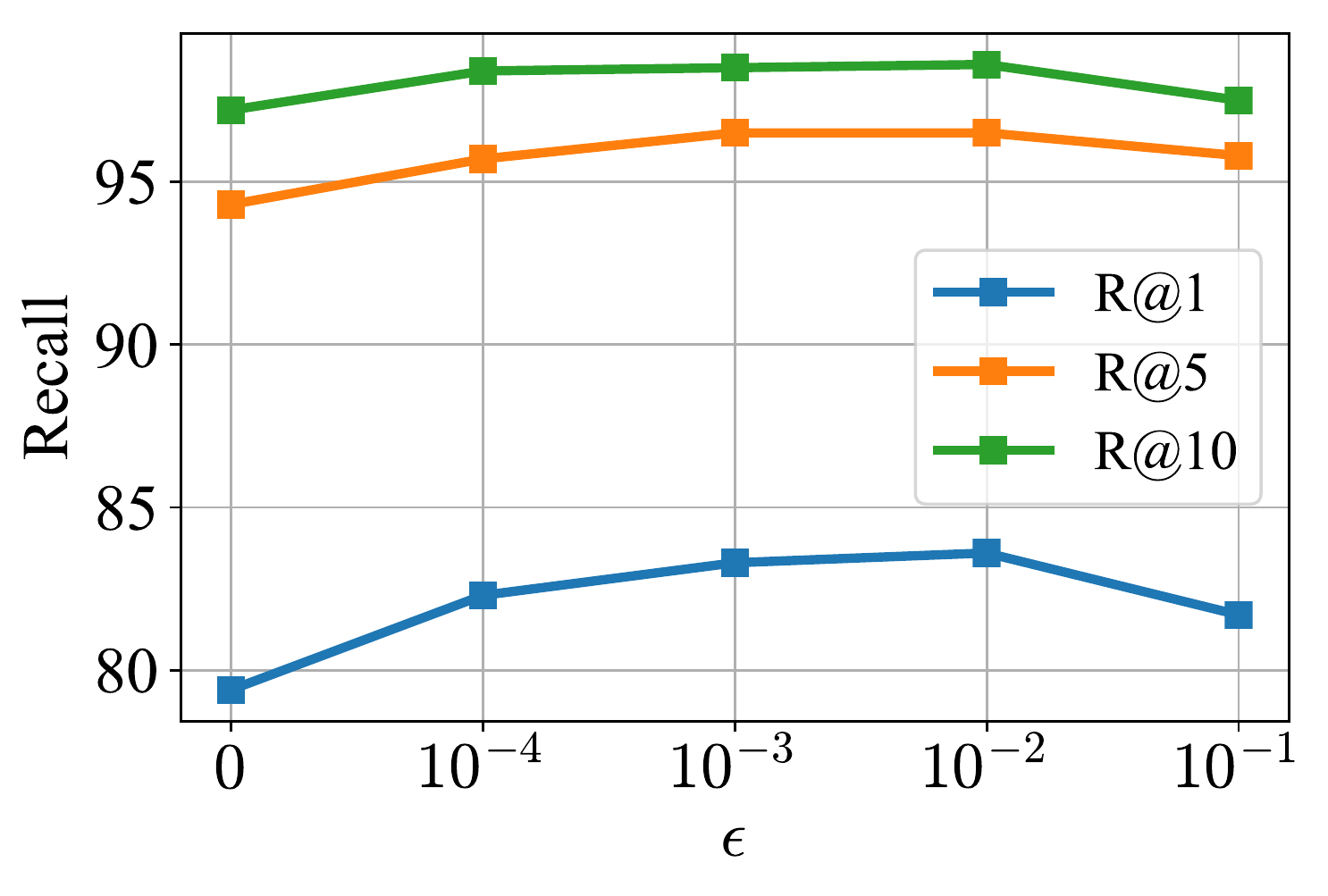}
		\caption{Image-to-Text}
		\label{epsilon_i2t}
	\end{subfigure}
	\hfill
	\begin{subfigure}{0.49\linewidth}
		\includegraphics[width=\linewidth]{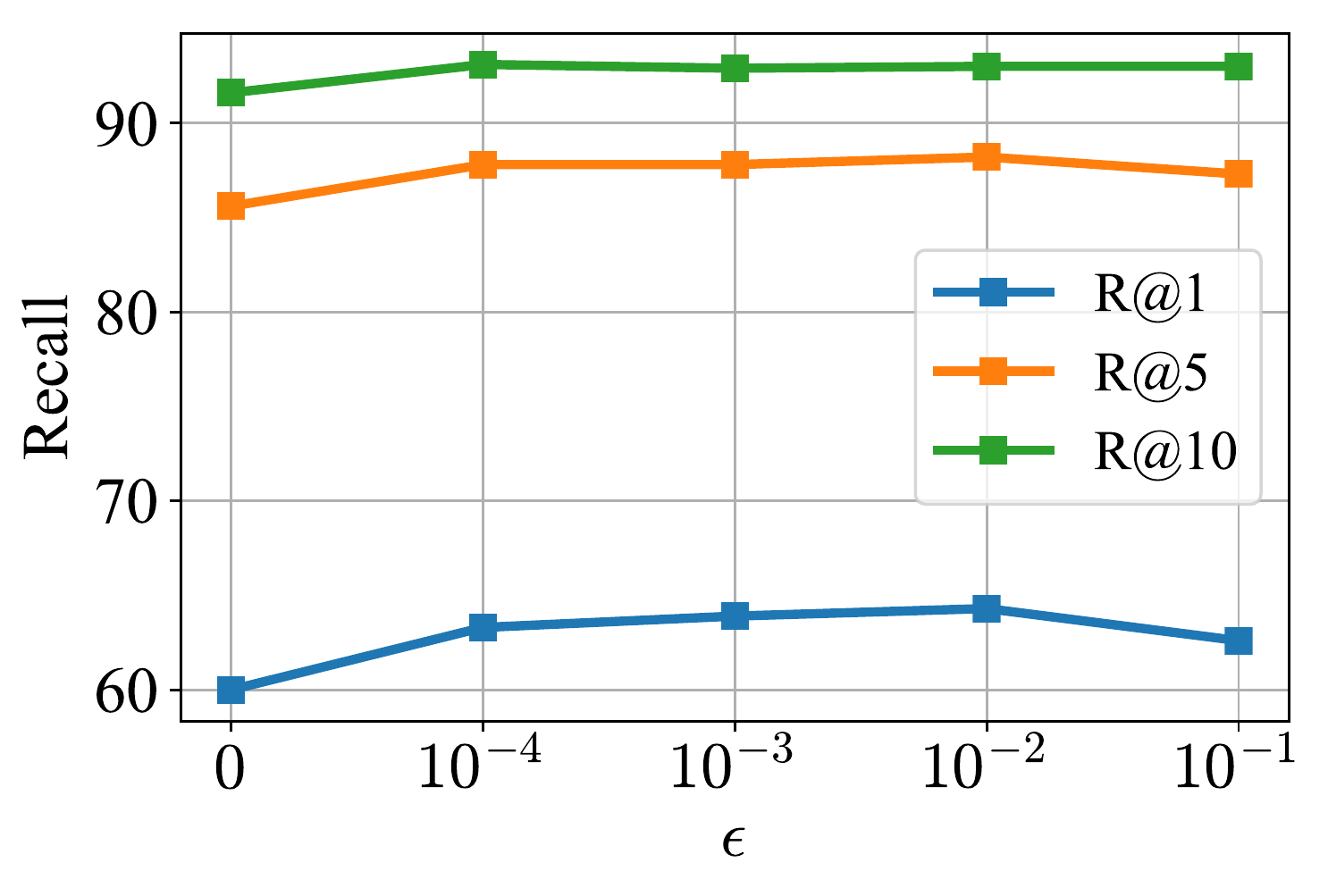}
		\caption{Text-to-Image}
		\label{epsilon_t2i}
	\end{subfigure}
	\vspace{-5pt}
	\caption{Effects of different configurations of the parameter on Flickr30K test set using RVSE (MLP) model.}
	\label{epsilon}
\end{figure}
\begin{figure}
	\centering
	\begin{subfigure}{0.49\linewidth}
		\includegraphics[width=\linewidth]{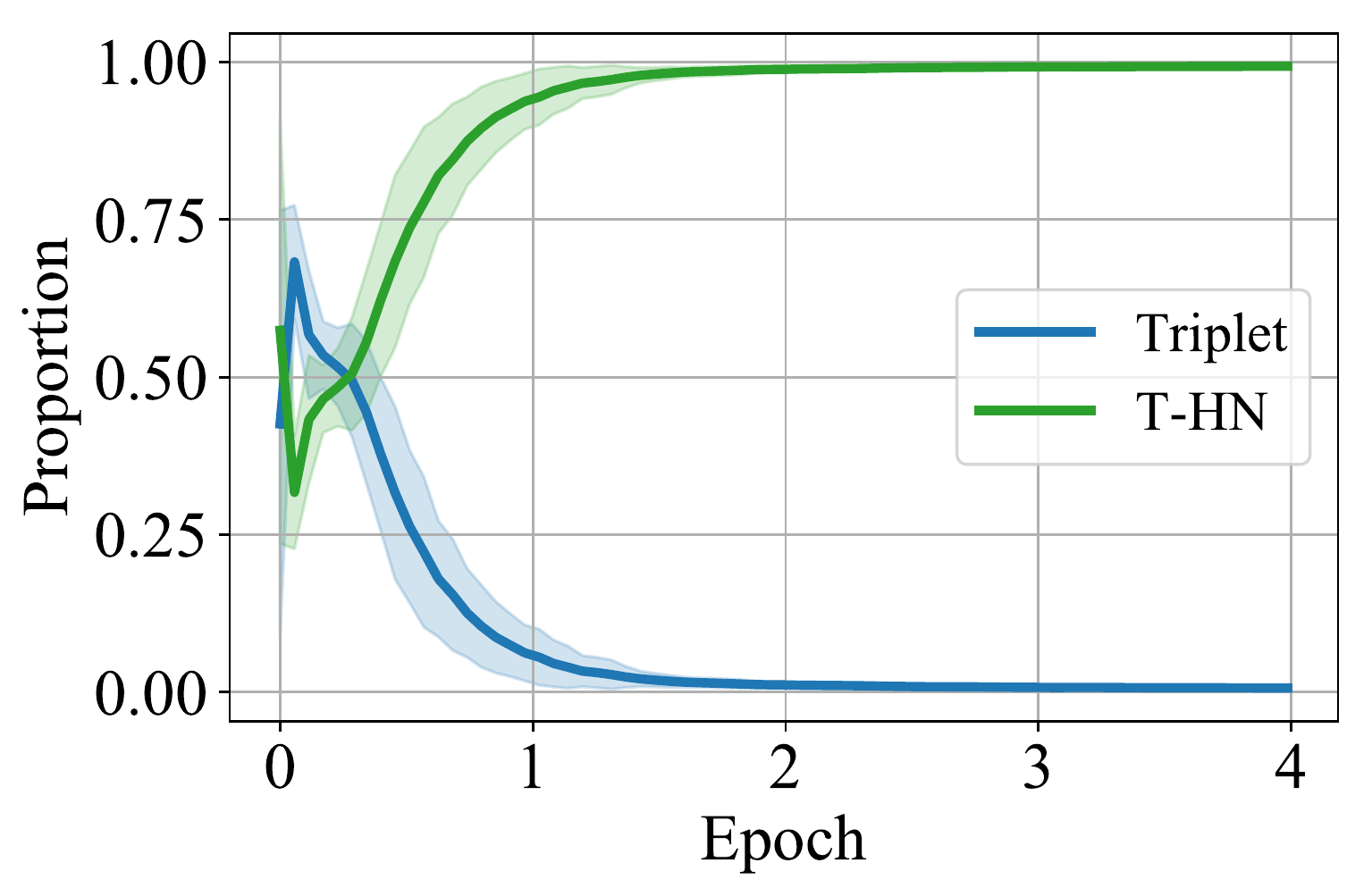}
		\caption{Flickr30K}
		\label{f30k_train_loss_selhn}
	\end{subfigure}
	\hfill
	\begin{subfigure}{0.49\linewidth}
		\includegraphics[width=\linewidth]{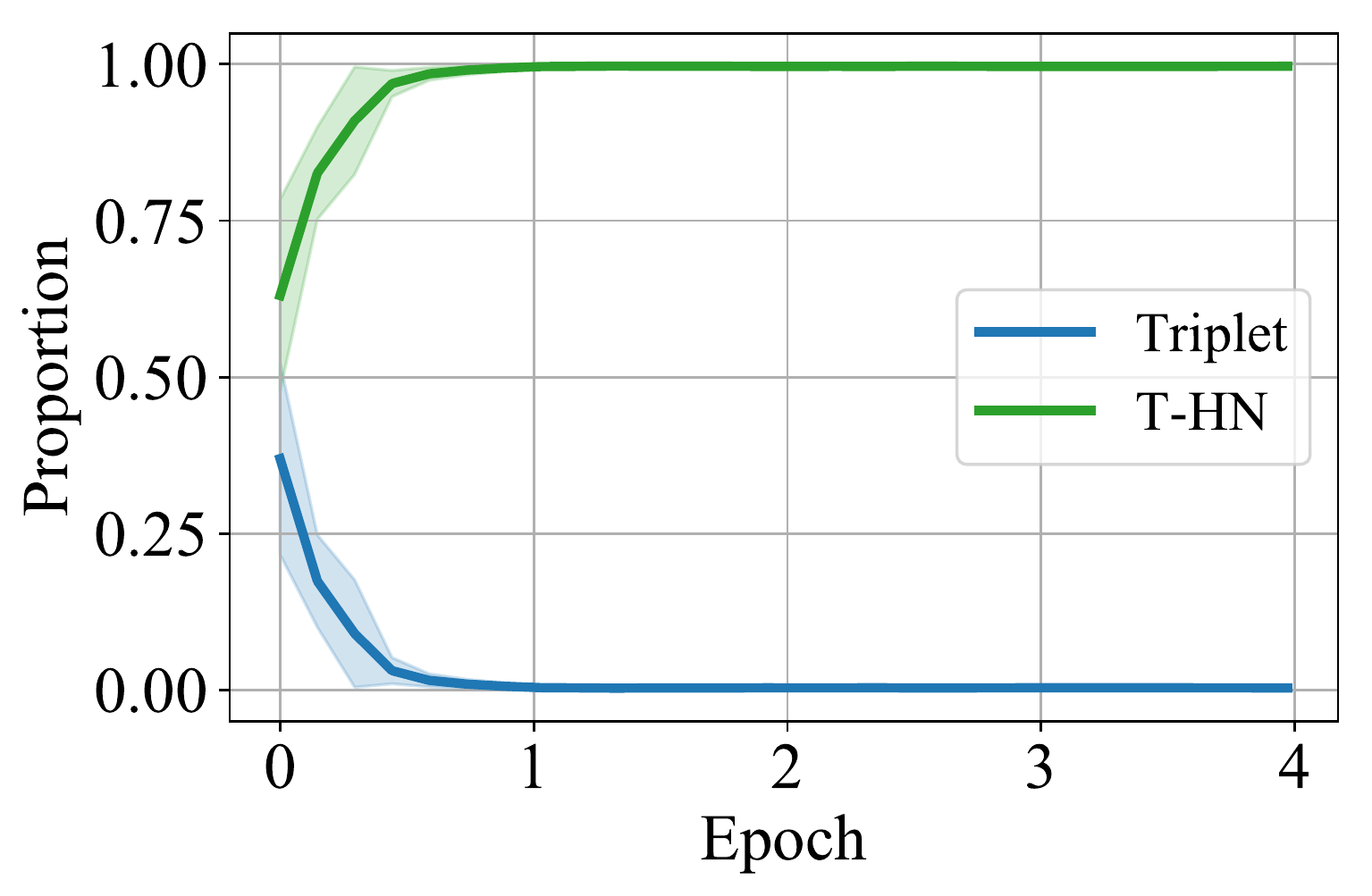}
		\caption{MS-COCO}
		\label{coco_train_loss_selhn}
	\end{subfigure}
	\vspace{-5pt}
	\caption{
		Loss function selection during training.
		The vertical axis is the proportion of $ \mathcal{L}_{\text{Triplet}} $ and $ \mathcal{L}_{\text{T-HN}} $ to $ \mathcal{L}_{\text{T-SelHN}} $.}
	\vspace{-10pt}
	\label{train_loss_selhn}
\end{figure}
There is one hyper-parameter threshold $ \epsilon $ in our SelHN that can be tuned. 
We experiment with several parameter settings on Flickr30K using the RVSE++ model.
$ \epsilon = 0 $ means the model is trained with $ \mathcal{L}_{\text{T-HN}} $.
As shown in \figurename~\ref{epsilon}, when $ \epsilon $ is in the range of $ 10^{-4} $ to $ 10^{-1} $, the matching performance of the model is better than when $ \epsilon = 0 $.
Overall, the matching performance of the ITM model does not change significantly with $ \epsilon $. 
This shows that our proposed SelHN is not sensitive to hyper-parameter and is convenient to apply to various ITM models.
In particular, in \figurename~\ref{epsilon_i2t}, when the value of $ \epsilon $ is $ 10^{-1} $, the matching performance decreases slightly.
Since when $ \epsilon $ is large, SelHN will discard more hard negative samples for optimization, resulting in a decrease in the matching performance of the model.

Our SelHN chooses whether to mine hard negative samples according to the gradient vanishing condition.
\figurename~\ref{train_loss_selhn} shows the choice of loss function during RVSE++ model training.
In the early stage of training, some triplets do not mine hard negative samples, since gradient vanishing is easy to occur at the beginning of training.
After training the model with triplet loss to increase $ \Delta s $, most triplets mine hard negative samples so that the model can learn more discriminative representations.

\subsection{Efficiency Analysis}
\begin{figure}
	\centering
	\includegraphics[width=0.7\linewidth]{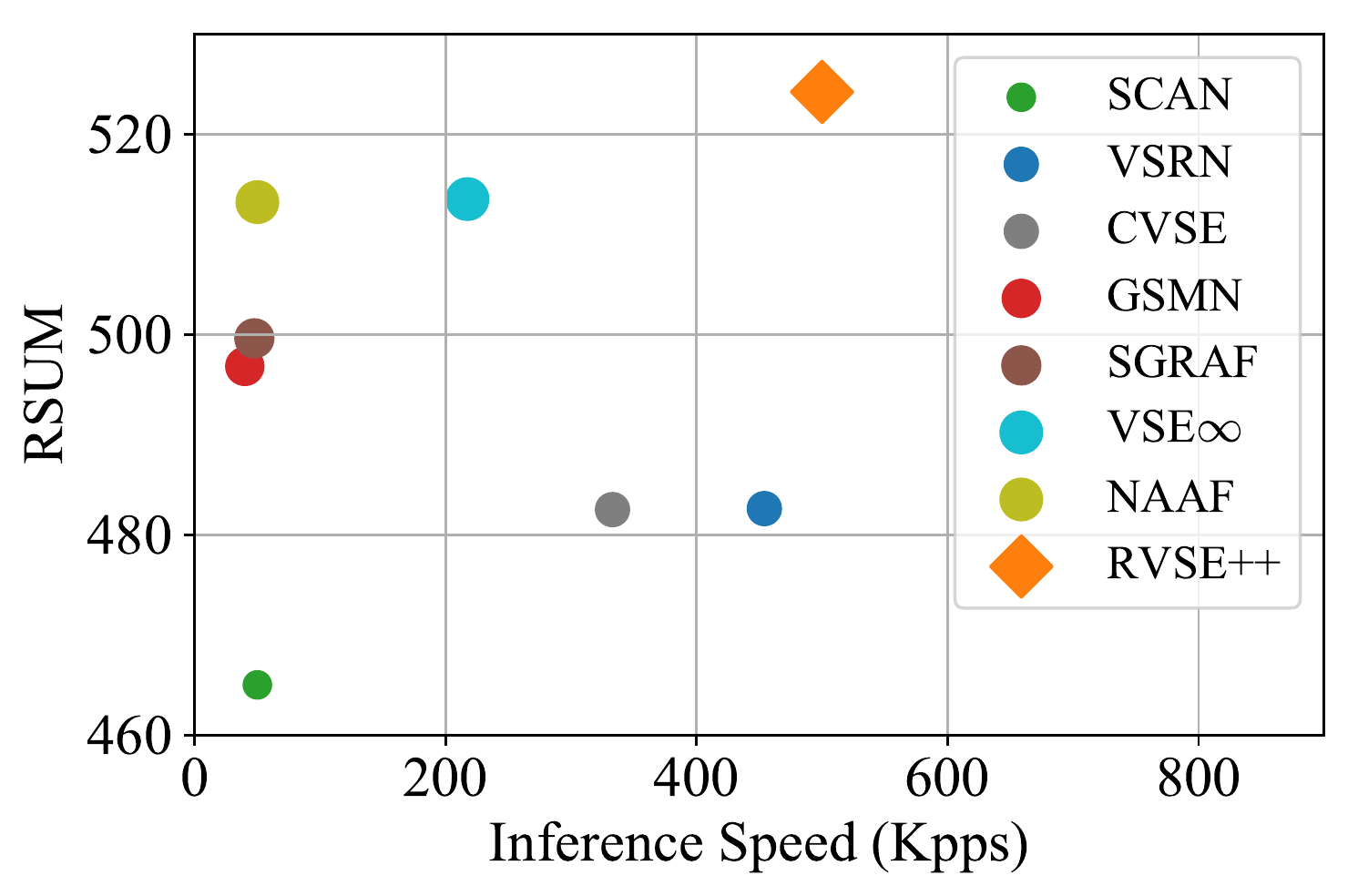}
	\vspace{-5pt}
	\caption{
		Comparison of matching performance and speed on Flickr30K test set.
		Kpps denote how many image-text pair matching score is computed per second.
	}
	\vspace{-10pt}
	\label{speed}
\end{figure}
The matching speed is also important in the real application scenario. 
Thus, we report both matching performance and speed for a more comprehensive comparison on Flickr30K.
We compare our RVSE++ with several state-of-the-art ITM methods. 
Note that their matching speed is reported by reimplementing their open-source codes in the same experimental setup.
As shown in \figurename~\ref{speed}, we can see that our RVSE++ leads other methods in both matching speed and matching performance. 
Therefore, our RVSE++ is superior to these methods from both perspectives of effectiveness and efficiency.

\section{Conclusion}
In this paper, we provide a solution to the gradient vanishing during ITM model training.
Specifically, we derive the condition under which the gradient vanishes during training. 
To alleviate the gradient vanishing, we propose a SelHN strategy, which chooses whether to mine hard negative samples according to the condition.
SelHN can be plug-and-play applied to existing ITM models to give them better training behavior.
To further ensure the back-propagation of gradients, we construct an RVSE++ model, which outperforms state-of-the-art methods by a large margin.
In future work, we plan to explore training performance on more vision-language cross-modal tasks.

\cleardoublepage
\appendix
In this appendix, we provide implementation details and experimental results omitted from the main text.


\section{Implementation Details}
Our all experiments are conducted on an NVIDIA GeForce RTX 3090 GPU using PyTorch.
For SelHN, hyper-parameters are set as $ \epsilon = 0.01 $ and $ \lambda = 0.2 $  for both datasets. 
VSE++ (FC), VSE (MLP) and RVSE (MLP) are trained using AdamW~\cite{loshchilov2018decoupled} for 20 epochs, with a batch size of 128 for both datasets.
The embedding dimension $ d $ is set to 1,024.
The learning rate of the model is set as 0.0005. 
RVSE++ is trained using AdamW for 30 epochs, with a batch size of 128 for both datasets. 
The embedding dimension $ d $ is set to 1,024.
The initial learning rate of the model is set as 0.0005 for first 15 epochs, and then decays by a factor of 10 for the last 15 epochs.

\begin{figure}
	\centering
	\includegraphics[width=0.3\linewidth]{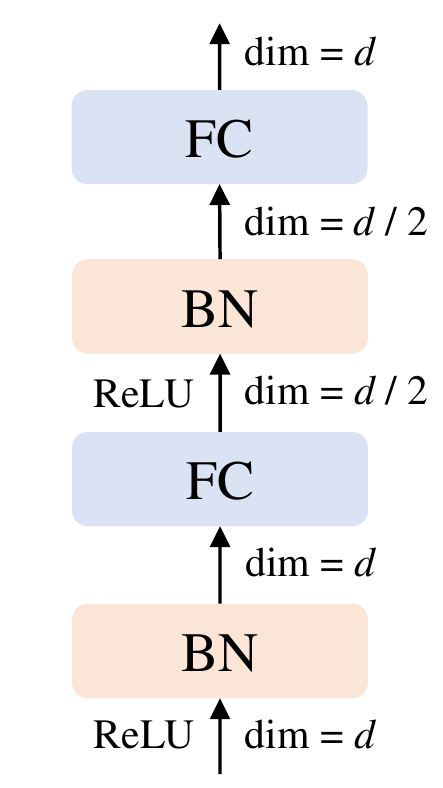}
	\vspace{-5pt}
	\caption{Network structure of MLP.}
	\label{mlp}
\end{figure}

The structure of two-layer MultiLayer Perceptron (MLP) used in the VSE (MLP), RVSE (MLP) and RVSE++ is shown in \figurename~\ref{mlp}.
The MLP is a bottleneck structure consisting of two Batch Normalization (BN) layers and two Fully Connected (FC) layers.
The input and output dimensions of the MLP are embedding dimension $ d $.
The hidden layer dimension of the MLP is $ d/2 $.

\section{Experimental Results}
\subsection{Improvements on Existing ITM Models}
To justify the superiority of our SelHN over the existing ITM models, we conduct experiments on SCAN~\cite{lee2018stacked}, BFAN~\cite{liu2019focus}, and SGRAF~\cite{diao2021similarity} by only replacing the loss functions.
SCAN and BFAN belong to the CA method, and SGRAF belongs to the combined method of VSE and CA.
Our previous experiments and gradient analysis are mainly based on the VSE model. 
This experiment shows that the conclusions we obtain can be extended to other types of models.
\tablename~\ref{exist_all} shows the improvements on existing ITM models on Flickr30K dataset.
\tablename~\ref{exist_all} reports the matching performance of both the single model and the ensemble model.
By replacing the loss function with our SelHN, the performance of the three baseline methods is improved.
Applying our SelHN can improve the matching performance of the model, whether on a single model or an ensemble model.

\begin{table}[t]
	\setlength\tabcolsep{2pt}
	\footnotesize
	\begin{center}
		\begin{tabular}{lcccccccccccc}
			\toprule[1pt]
			Eval Task $ \rightarrow $
			& \multicolumn{3}{c}{Image-to-Text} & \multicolumn{3}{c}{Text-to-Image} & \multirow{2}*{RSUM}\\
			\cline{1-7}
			\specialrule{0em}{2pt}{0pt}
			Method $ \downarrow $ & R@1 & R@5 & R@10 & R@1 & R@5 & R@10 \\
			\hline
			
			\specialrule{0em}{2pt}{0pt}
			SCAN I2T AVG~\cite{lee2018stacked} & 67.9 & 89.0 & 94.4 & 43.9 & 74.2 & 82.8 & 452.2 \\
			\rowcolor{black!10}
			+ $\mathcal{L}_{\text{T-SelHN}}$ & 66.3 & 90.2 & 95.2 & 45.5 & 75.1 & 83.7 & \textbf{456.0} \\
			\hline
			\specialrule{0em}{2pt}{0pt}
			SCAN T2I AVG~\cite{lee2018stacked} & 61.8 & 87.5 & 93.7 & 45.8 & 74.4 & 83.0 & 446.2 \\
			\rowcolor{black!10}
			+ $\mathcal{L}_{\text{T-SelHN}}$ & 70.3 & 93.1 & 95.7 & 53.3 & 79.0 & 86.0 & \textbf{477.5} \\ 
			\hline
			\specialrule{0em}{2pt}{0pt}
			SCAN$ _{\text{(\textit{ECCV}'18)}}$~\cite{lee2018stacked} & 67.4 & 90.3 & 95.8 & 48.6 & 77.7 & 85.2 & 465.0 \\
			\rowcolor{black!10}
			+ SelHN & 73.4 & 93.7 & 96.6 & 53.5 & 81.1 & 87.9 & \textbf{486.2} \\
			\hline
			
			\specialrule{0em}{2pt}{0pt}
			BFAN (equal)~\cite{liu2019focus} & 64.5 & 89.7 & - & 48.8 & 77.3 & - & - \\
			\rowcolor{black!10}
			+ $\mathcal{L}_{\text{T-SelHN}}$ & 71.5 & 92.5 & 96.0 & 53.5 & 78.6 & 85.7 & \textbf{477.8} \\
			\hline
			\specialrule{0em}{2pt}{0pt}
			BFAN (prob)~\cite{liu2019focus} & 65.5 & 89.4 & - & 47.9 & 77.6 & - & - \\
			\rowcolor{black!10}
			+ $\mathcal{L}_{\text{T-SelHN}}$ & 71.8 & 92.3 & 95.9 & 52.5 & 78.8 & 85.6 & \textbf{476.8} \\
			\hline
			\specialrule{0em}{2pt}{0pt}
			BFAN$ _{\text{(\textit{MM}'19)}}$~\cite{liu2019focus} & 68.1 & 91.4 & 95.9 & 50.8 & 78.4 & 85.8 & 470.4 \\
			\rowcolor{black!10}
			+ SelHN & 75.3 & 93.4 & 97.1 & 55.2 & 80.9 & 87.3 & \textbf{489.1} \\
			\hline
			
			\specialrule{0em}{2pt}{0pt}
			SAF~\cite{diao2021similarity} & 73.7 & 93.3 & 96.3 & 56.1 & 81.5 & 88.0 & 488.9 \\
			\rowcolor{black!10}
			+ $\mathcal{L}_{\text{T-SelHN}}$ & 76.0 & 94.8 & 97.1 & 57.9 & 83.5 & 89.2 & \textbf{498.5} \\
			\hline
			\specialrule{0em}{2pt}{0pt}
			SGR~\cite{diao2021similarity} &  75.2 & 93.3 & 96.6 & 56.2 & 81.0 & 86.5 & 488.8 \\
			\rowcolor{black!10}
			+ $\mathcal{L}_{\text{T-SelHN}}$ & 77.8 & 94.6 & 98.0 & 58.2 & 82.2 & 87.2 & \textbf{498.0} \\
			\hline
			\specialrule{0em}{2pt}{0pt}
			SGRAF$ _{\text{(\textit{AAAI}'21)}}$~\cite{diao2021similarity} & 77.8 & 94.1 & 97.4 & 58.5 & 83.0 & 88.8 & 499.6 \\
			\rowcolor{black!10}
			+ $ \mathcal{L}_{\text{T-SelHN}} $ & 80.0 & 95.1 & 98.2 & 59.8 & 84.5 & 89.5 & \textbf{507.1} \\
			
			\bottomrule[1pt]
		\end{tabular}
	\end{center}
	\vspace{-15pt}
	\caption{Matching performance of applying SelHN to existing ITM models on Flickr30K.}
	\label{exist_all}
\end{table}
\tablename~\ref{exist_coco1k} and \tablename~\ref{exist_coco5k} shows the improvements on existing ITM models on MS-COCO 1K and 5K test set.
After applying our SelHN, BFAN and SGRAF have improved on most of the evaluation metrics.
Our SelHN strategy can be plug-and-play applied to existing ITM models. 
The experimental results show that the application of SelHN can give full play to the representation ability of the model and achieve better matching performance.
\begin{table}[t]
	\setlength\tabcolsep{2pt}
	\footnotesize
	\begin{center}
		\begin{tabular}{lcccccccccccc}
			\toprule[1pt]
			Eval Task $ \rightarrow $
			& \multicolumn{3}{c}{Image-to-Text} & \multicolumn{3}{c}{Text-to-Image} & \multirow{2}*{RSUM}\\
			\cline{1-7}
			\specialrule{0em}{2pt}{0pt}
			Method $ \downarrow $ & R@1 & R@5 & R@10 & R@1 & R@5 & R@10 \\
			\hline
			
			\specialrule{0em}{2pt}{0pt}
			BFAN$ _{\text{(\textit{MM}'19)}}$~\cite{liu2019focus} & 74.9 & 95.2 & 98.3 & 59.4 & 88.4 & 94.5 & 510.7 \\
			\rowcolor{black!10}
			+ $ \mathcal{L}_{\text{T-SelHN}} $ & 76.2 & 95.6 & 98.4 & 60.7 & 88.0 & 94.4 & \textbf{513.4} \\
			\hline
			
			SGRAF$ _{\text{(\textit{AAAI}'21)}}$~\cite{diao2021similarity} & 79.6 & 96.2 & 98.5 & 63.2 & 90.7 & 96.1 & 524.3 \\
			\rowcolor{black!10}
			+ $ \mathcal{L}_{\text{T-SelHN}} $ & 79.9 & 96.4 & 98.6 & 65.2 & 90.9 & 95.8 & \textbf{526.9} \\
			
			\bottomrule[1pt]
		\end{tabular}
	\end{center}
	\vspace{-15pt}
	\caption{Matching performance of applying SelHN to existing ITM models on MS-COCO 1K test set.}
	\label{exist_coco1k}
\end{table}
\begin{table}[t]
	\setlength\tabcolsep{2pt}
	\footnotesize
	\begin{center}
		\begin{tabular}{lcccccccccccc}
			\toprule[1pt]
			Eval Task $ \rightarrow $
			& \multicolumn{3}{c}{Image-to-Text} & \multicolumn{3}{c}{Text-to-Image} & \multirow{2}*{RSUM}\\
			\cline{1-7}
			\specialrule{0em}{2pt}{0pt}
			Method $ \downarrow $ & R@1 & R@5 & R@10 & R@1 & R@5 & R@10 \\
			\hline
			
			\specialrule{0em}{2pt}{0pt}
			BFAN$ _{\text{(\textit{MM}'19)}}$~\cite{liu2019focus} & 52.9 & 82.8 & 90.6 & 38.3 & 67.8 & 79.3 & 411.7 \\
			\rowcolor{black!10}
			+ $ \mathcal{L}_{\text{T-SelHN}} $ & 55.6 & 82.2 & 90.6 & 39.3 & 68.0 & 78.6 & \textbf{414.3} \\
			\hline
			
			SGRAF$ _{\text{(\textit{AAAI}'21)}}$~\cite{diao2021similarity} & 57.8 & - & 91.6 & 41.9 & - & 81.3 & - \\
			\rowcolor{black!10}
			+ $ \mathcal{L}_{\text{T-SelHN}} $ & 59.8 & 85.5 & 92.4 & 43.8 & 72.6 & 82.7 & \textbf{436.8} \\
			
			\bottomrule[1pt]
		\end{tabular}
	\end{center}
	\vspace{-15pt}
	\caption{Matching performance of applying SelHN to existing ITM models on MS-COCO 5K test set.}
	\label{exist_coco5k}
\end{table}

\subsection{Comparisons with the Other Loss Functions}
Our SelHN focuses on alleviating gradient vanishing in ITM model training. 
Therefore, we compare SelHN with other methods that improve training behavior.

\textbf{Semi-Hard Negative mining (SHN)}~\cite{schroff2015facenet} does not optimize hard negative samples, and only mines semi-hard negative samples with $ s_{n}<s_{p} $ for optimization.
The triplet loss with SHN applied to ITM contains both image-to-text and text-to-image directions and can be expressed as:
\begin{equation} \label{SHN}
	\begin{aligned}
		\mathcal{L}_{\text{T-SHN}}
		= \sum_{i=1}^{B}
		\left(
		[ 
		s(V_{i}, \hat{T}_{i}) - s(V_{i}, T_{i}) + \lambda 
		]_{+}
		\right. \\
		\left. 
		+ [ 
		s(\hat{V}_{i}, T_{i}) - s(V_{i}, T_{i}) + \lambda 
		]_{+}
		\right),
	\end{aligned}
\end{equation}
where
\begin{equation} \label{SHN}
	\begin{aligned}
		\hat{T}_{i} 
		= \arg\max_{j = 1, 
			s(V_{i}, T_{j}) 
			< s(V_{i}, T_{i})
		}^{B} 
		s{(V_{i}, T_{j})} \\
		\hat{V}_{i} 
		= \arg\max_{j = 1,
			s(V_{j}, T_{i}) 
			< s(V_{i}, T_{i})
		}^{B} 
		s{(V_{j}, T_{i})},
	\end{aligned}
\end{equation}
are the semi-hard negative samples. 

\textbf{Selectively Contrastive Triplet loss (SCT)}~\cite{xuan2020hard} uses the contrastive loss to optimize hard negative samples and uses the triplet loss to optimize the remaining samples.
With $ V_{i} $ as the anchor, the SCT takes the form of:
\begin{equation}
	\mathcal{L}_{\text{SCT}}
	(V_{i}) 
	=
	\begin{cases}
		\mathcal{L}_{\text{T-HN}}
		(V_{i}) ,
		& s(V_{i}, \hat{T}_{i}) < s(V_{i}, T_{i}), \\
		s(V_{i}, \hat{T}_{i}), 
		& \text{otherwise},
	\end{cases}
\end{equation}
where
\begin{equation}
	\mathcal{L}_{\text{T-HN}}(V_{i})
	=
	[ 
	s(V_{i}, \hat{T}_{i}) - s(V_{i}, T_{i}) + \lambda 
	]_{+}.
\end{equation}

\begin{table}[t] 
	\setlength\tabcolsep{4pt}
	\footnotesize
	\begin{center}
		\begin{tabular}{lcccccccccccccccccccc}
			\toprule[1pt]
			Eval Task $ \rightarrow $
			& \multicolumn{3}{c}{Image-to-Text} & \multicolumn{3}{c}{Text-to-Image} & \multirow{2}*{RSUM} \\
			\specialrule{0em}{2pt}{0pt}
			Loss $ \downarrow $ & R@1 & R@5 & R@10 & R@1 & R@5 & R@10 \\
			\hline
			
			\specialrule{0em}{2pt}{0pt}
			Triplet & 64.5 & 89.5 & 95.0 & 48.4 & 77.4 & 86.0 & 460.7 \\
			HN & 76.0 & 92.5 & 97.0 & 53.9 & 81.2 & 88.2 & 488.8 \\
			SHN & 74.0 & 93.4 & 97.4 & 55.3 & 82.2 & 88.8 & 491.0 \\
			SCT & 75.8 & 93.6 & 97.1 & 55.3 & 82.0 & 88.6 & 492.3 \\
			\rowcolor{black!10}
			$ \mathcal{L}_{\text{T-SelHN}} $ & 77.1 & 93.6 & 97.1 & 56.3 & 82.8 & 89.2 & \textbf{496.1} \\
			
			\bottomrule[1pt]
		\end{tabular}
	\end{center}
	\vspace{-15pt}
	\caption{Performance comparison with other methods that improve training behavior on MS-COCO 1K test set.}
	\label{other_triplet_coco}
\end{table}
The experimental results of SelHN compared to the above losses on the VSE (FC) model on MS-COCO 1K test set are shown in \tablename~\ref{other_triplet_coco}.
Compared with these loss functions, SelHN improves most evaluation metrics.
SHN and SCT improve the training behavior. 
But they abandon optimization for triplet consisting of hard negative samples.
Hard negative samples are crucial for learning a discriminative model. 
SelHN mines hard negative samples when gradient vanishing is unlikely to occur. 
Therefore, SelHN not only improves the training behavior but also gives full play to the role of hard negative mining.

\begin{table}[t] 
	\setlength\tabcolsep{2pt}
	\footnotesize
	\begin{center}
		\begin{tabular}{lcccccccccccc}
			\toprule[1pt]
			Eval Task $ \rightarrow $
			& \multicolumn{3}{c}{Image-to-Text} & \multicolumn{3}{c}{Text-to-Image} & \multirow{2}*{RSUM}\\
			\cline{1-7}
			\specialrule{0em}{2pt}{0pt}
			Method $ \downarrow $ & R@1 & R@5 & R@10 & R@1 & R@5 & R@10 \\
			\hline
			\specialrule{0em}{2pt}{0pt}
			SCAN$_{\text{(\textit{ECCV}'18)}}$~\cite{lee2018stacked} & 50.4 & 82.2 & 90.0 & 38.6 & 69.3 & 80.4 & 410.9 \\
			BFAN$ _{\text{(\textit{MM}'19)}} $~\cite{liu2019focus} & 52.9 & 82.8 & 90.6 & 38.3 & 67.8 & 79.3 & 411.7 \\
			VSRN$_{\text{(\textit{ICCV}'19)}}$~\cite{li2019visual} & 53.0 & 81.1 & 89.4 & 40.5 & 70.6 & 81.1 & 415.7 \\
			IMRAM$_{\text{(\textit{CVPR}'20)}}$~\cite{chen2020imram} & 53.7 & 83.2 & 91.0 & 39.7 & 69.1 & 79.8 & 416.5 \\
			SGRAF$_{\text{(\textit{AAAI}'21)}}$~\cite{diao2021similarity} & 57.8 & - & 91.6 & 41.9 & - & 81.3 & - \\
			VSE$\infty$ $ _{\text{(\textit{CVPR}'21)}} $~\cite{chen2021learning} & 58.3 & 85.3 & 92.3 & 42.4 & 72.7 & 83.2 & 434.3 \\
			CMCAN$_{\text{(\textit{AAAI}'22)}}$~\cite{zhang2022show} & 61.5 & - & 92.9 & 44.0 & - & 82.6 & - \\
			NAAF$_{\text{(\textit{CVPR}'22)}}$~\cite{zhang2022negative} & 58.9 & 85.2 & 92.0 & 42.5 & 70.9 & 81.4 & 430.9 \\
			\hline
			
			\specialrule{0em}{2pt}{0pt}
			\rowcolor{black!10}
			\textbf{RVSE++} & 60.6 & 86.4 & 92.8 & 44.5 & 74.5 & 84.5 & \textbf{443.4} \\
			\bottomrule[1pt]
		\end{tabular}
	\end{center}
	\vspace{-15pt}
	\caption{Performance comparison with the state-of-the-art ITM methods on MS-COCO 5K test set.}
	\vspace{-10pt}
	\label{coco5k}
\end{table}
\subsection{Comparisons with state-of-the-art Methods}
The experimental results on the MS-COCO 5K test set is shown in \tablename~\ref{coco5k} . 
We can see that our RVSE++ outperforms these methods in terms of most evaluation metrics. 
Compared with NAAF, RVSE++ surpasses all its evaluation performance, with relative 12.5\% on RSUM.
Experimental results demonstrate the effectiveness of our RVSE++. 
The residual design of the RVSE++ model and the guarantee of the SelHN strategy for gradient backpropagation enable the RVSE++ model to give full play to the representation ability of the model.

\cleardoublepage
{\small
\bibliographystyle{ieee_fullname}
\bibliography{egbib}
}

\end{document}